\pdfoutput=1

\documentclass[11pt]{article}
\usepackage[table]{xcolor}
\usepackage[final]{acl}

\usepackage{xcolor}
\usepackage{times}
\usepackage{latexsym}
\usepackage{graphicx}
\usepackage{float}
\usepackage{colortbl}
\usepackage{multirow}
\usepackage{booktabs}
\usepackage{caption}
\usepackage{subcaption}
\usepackage{rotating}
\usepackage[export]{adjustbox}
\usepackage{dirtytalk} 

\usepackage{array}      
\usepackage{booktabs}   
\usepackage{graphicx}   
\usepackage{tcolorbox}
\definecolor{D55E00}{HTML}{D55E00}
\definecolor{009E73}{HTML}{009E73}
\definecolor{0072B2}{HTML}{0072B2}
\usepackage{tikz}
\newcommand{\tikzxmark}{%
\tikz[scale=0.23] {
    \draw[line width=0.7,line cap=round] (0,0) to [bend left=6] (1,1);
    \draw[line width=0.7,line cap=round] (0.2,0.95) to [bend right=3] (0.8,0.05);
}}

\usepackage{amsmath}
\usepackage{amssymb}

\usepackage{booktabs}
\usepackage[table]{xcolor}
\usepackage{siunitx}
\usepackage{caption}
\usepackage{graphicx}

\definecolor{Best}{HTML}{D9EAD3}

\usepackage{algorithm}
\usepackage{algorithmicx}   
\usepackage{algpseudocode}
\usepackage{float}
\usepackage{pdflscape}
\usepackage{hyperref}
\usepackage[]{acronym}
\usepackage{amsmath}

\usepackage{amssymb}

\newtheorem{proof}{Proof}

\usepackage{comment}
\usepackage[T1]{fontenc}

\usepackage[utf8]{inputenc}

\usepackage{microtype}

\usepackage{inconsolata}

\usepackage{graphicx}
\usepackage{enumitem}
\usepackage{makecell}
\title{GUARD: Glocal Uncertainty-Aware Robust Decoding for Effective \\ and Efficient Open-Ended Text Generation}

\author{
  Yuanhao Ding\thanks{\ \ Equal contribution}$^{1}$~~~
  Esteban Garces Arias\footnotemark[1]$^{2,3}$~~~
  Meimingwei Li\footnotemark[1]$^{2}$~~~
  Julian Rodemann$^{2,4}$\\
  \textbf{Matthias Aßenmacher}$^{2,3}$~~~
  \textbf{Danlu Chen}$^{5}$~~~
  \textbf{Gaojuan Fan}$^{1}$~~~
  \textbf{Christian Heumann}$^{2}$\\
  \textbf{Chongsheng Zhang}\thanks{\ \ Corresponding author}$^{1}$ 
\\
  $^1$Henan University, $^2$Department of Statistics, LMU Munich\\
  $^3$Munich Center for Machine Learning (MCML)\\
  $^4$CISPA Helmholtz Center for Information Security, Saarbrücken\\
  $^5$University of California, San Diego\\ 
   \{yhding, fangaojuan, cszhang\}@henu.edu.cn\\ \{esteban.garcesarias, matthias\}@stat.uni-muenchen.de,  M.Li@campus.lmu.de\\ \{J.Rodemann, Christian.Heumann\}@lmu.de, danlu@ucsd.edu
}

\begin{document}
\maketitle

\begin{abstract}

Open-ended text generation faces a critical challenge: balancing coherence with diversity in LLM outputs. While contrastive search-based decoding strategies have emerged to address this trade-off, their practical utility is often limited by hyperparameter dependence and high computational costs. We introduce GUARD, a self-adaptive decoding method that effectively balances these competing objectives through a novel "Glocal" uncertainty-driven framework. GUARD combines global entropy estimates with local entropy deviations to integrate both long-term and short-term uncertainty signals. We demonstrate that our proposed global entropy formulation effectively mitigates abrupt variations in uncertainty, such as sudden overconfidence or high entropy spikes, and provides theoretical guarantees of unbiasedness and consistency. To reduce computational overhead, we incorporate a simple yet effective token-count-based penalty into GUARD. Experimental results demonstrate that GUARD achieves a good balance between text diversity and coherence, while exhibiting substantial improvements in generation speed. In a more nuanced comparison study across different dimensions of text quality, both human and LLM evaluators validated its remarkable performance. Our code is available at \url{https://github.com/YecanLee/GUARD}.


\end{abstract}

\section{Introduction}
\label{sec:introduction}

Neural text generation models based on the Transformer decoder \citep{vaswani2017attention} have revolutionized natural language generation tasks such as story creation \citep{storygeneration}, context completion \citep{contextcompletion}, or dialogue generation \citep{dustysteam}. Decoding strategies play a crucial role in text generation with large language models (LLMs), affecting both the quality of the generated text and the computational efficiency during inference. The most commonly used strategies can be broadly categorized in deterministic \cite{Freitag_2017,carlsson2024hyperfittingphenomenonsharpeningstabilizing} and stochastic approaches \cite{Fantopk,nguyen2024turningheatminpsampling,aichberger2024semanticallydiverselanguagegeneration}. Deterministic approaches are likelihood-based techniques that typically (over-)emphasize coherence at the cost of diversity and are prone to producing repetitive text, a phenomenon referred to as \textit{degeneration}. Stochastic approaches, on the other hand, aim to improve the diversity of texts, potentially sacrificing coherence and leading to semantic inconsistencies \citep{semanticinconsistency}. 

This inherent \textbf{coherence vs. diversity trade-off} has motivated recent work to develop decoding strategies able to balance these two competing objectives. In particular, Contrastive Search \citep[CS;][]{su2022contrastive} introduced a weighted combination of model confidence (favoring coherence) and a degeneration penalty (promoting diversity) controlled by fixed, tunable hyperparameters throughout the generation. Adaptive Contrastive Search \citep[ACS;][]{estebanACS} addressed this hyperparameter dependence by dynamically adjusting them based on the local uncertainty of the model. In doing so, this approach still has its limitations as it relies only on local uncertainty information, and it substantially increases the computational overhead at inference time. In this study, we aim to address the following research questions (RQ):

\begin{enumerate}
    \item \textbf{RQ1:} Can we improve text quality in open-ended text generation compared to existing approaches (\S\ref{sec:results})?
    \item \textbf{RQ2:} Can we smooth out abrupt entropy deviations during generation without compromising statistical properties such as unbiasedness and consistency (\S\ref{subsec:global-entropy})?
    \item \textbf{RQ3:} Can we substantially reduce the latency and computational cost compared to (A)CS (\S\ref{sec:results}, "Generation Speed")?
\end{enumerate}

To address the above questions, we introduce a novel concept called "\textbf{Glocal}" uncertainty, which combines \textit{global} entropy estimates with \textit{local} entropy deviations to integrate long-term and short-term uncertainty signals, thereby maintaining robust decoding while ensuring statistical unbiasedness and consistency. Based on Glocal uncertainty, we propose a novel decoding strategy, \textbf{Glocal Uncertainty-Aware Robust Decoding (GUARD)}, which effectively balances the coherence and diversity of the generated texts (\textbf{RQ1}), addresses the smoothness of entropy estimates (\textbf{RQ2}), and improves inference speed (\textbf{RQ3}).\\

\noindent \textbf{Our Contributions} can be summarized as follows: 

\begin{enumerate}[noitemsep,topsep=3pt]
    \item We introduce "Glocal" uncertainty, which integrates global entropy and instantaneous uncertainty signals to smooth out strong deviations in entropy, resulting in more robust decoding. We provide statistical guarantees (unbiasedness and consistency) for the proposed global entropy estimator, which is an integral part of the Glocal uncertainty.
    \item  Based on Glocal uncertainty, we design GUARD, a self-adaptive decoding strategy that effectively and efficiently balances coherence and diversity. Compared to (A)CS, it reduces the computational cost remarkably by incorporating a simple, yet effective, token count penalty.
    \item We evaluate GUARD through comprehensive experimentation that spans diverse models, datasets, and decoding configurations, comparing it against state-of-the-art (SOTA) decoding strategies. Our assessment employs automatic, human, and LLM evaluations, which consistently show that GUARD represents a high-performing decoding strategy. 
\end{enumerate}

\section{Related work}
\label{sec:related_work}

\paragraph{Deterministic Strategies} primarily rely on maximizing the sequence probability. Greedy Search selects tokens based on pure maximum likelihood, while Beam search \cite{Freitag_2017} improves upon this by maintaining multiple candidates simultaneously, ultimately yielding the most probable sequence. Further research \citep{Ashwindecode, Holtzmandecode, garces-arias-etal-2025-decoding} has demonstrated that this often results in text degeneration and repetitive patterns. Recent developments, such as \textit{hyperfitting} \cite{carlsson2024hyperfittingphenomenonsharpeningstabilizing}, show promising solutions to this problem, even within greedy search frameworks.

\begin{figure*}[!ht]
\centering
\includegraphics[width=\textwidth]
{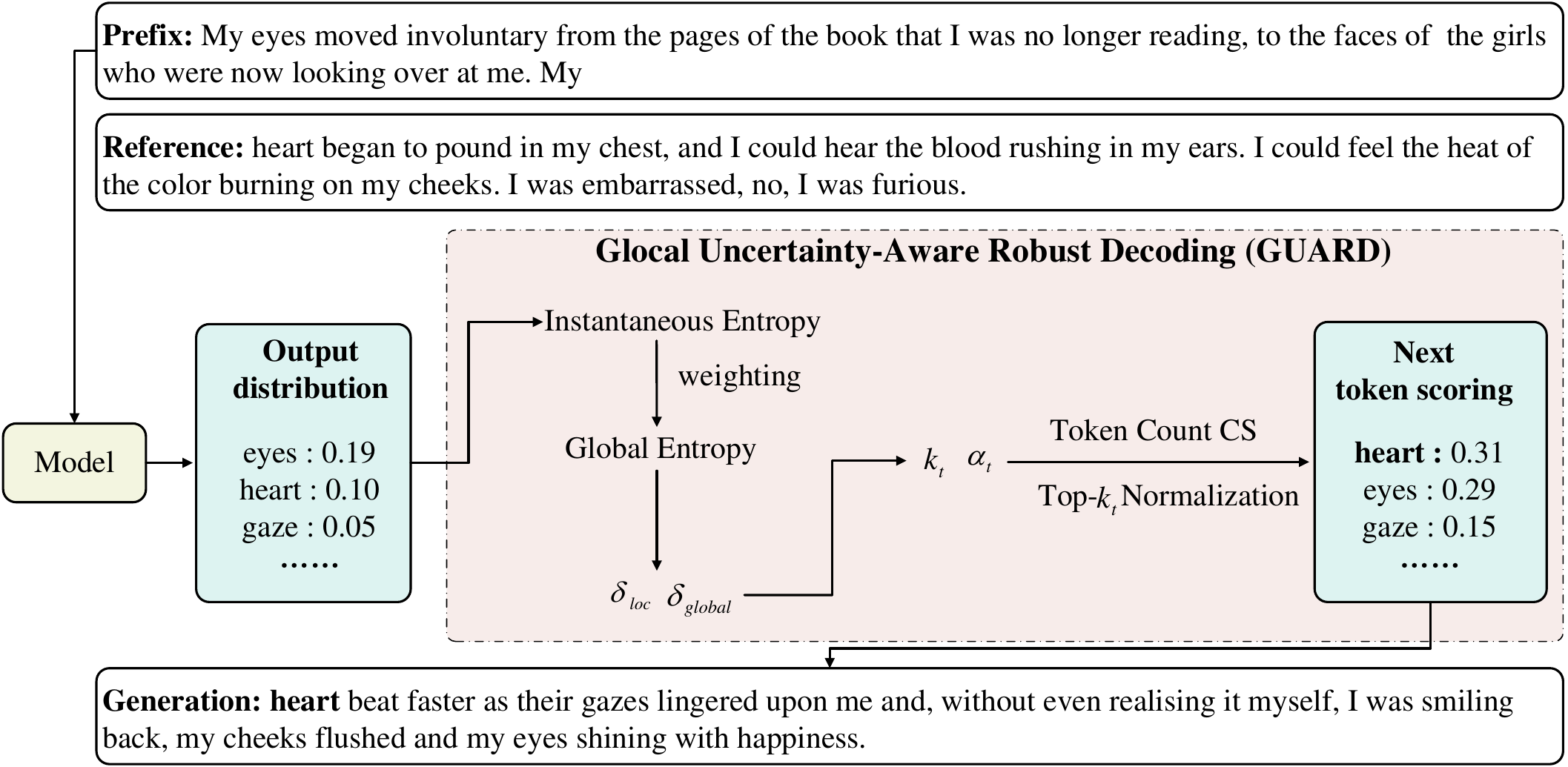}
\caption{GUARD leverages local and global Shannon entropy deviations as proxies for model uncertainty, automatically adjusting CS parameters and incorporating a token count penalty for next-token selection.}
\label{fig:figure_two}
\end{figure*}

\paragraph{Stochastic Strategies} based on sampling have emerged as a response to the limitations of deterministic approaches, particularly addressing text monotony and degeneration. Temperature sampling \cite{ackley1985learning} introduces a hyperparameter $\tau$ that modulates the sharpness of the output distribution, enabling control over generation diversity. Top-$k$ \citep{Fantopk} and Top-$p$ sampling \citet{Holtzmandecode} truncate the output distribution to avoid sampling from less reliable tail probabilities. Typical sampling \cite{typicalsampling} constrains the sampling distribution to tokens whose negative log-probabilities fall within a specific range of the model's conditional entropy, while min-$p$ sampling \cite{nguyen2024turningheatminpsampling} introduces dynamic truncation based on the model's confidence. Adaptive Decoding \citep[AD,][]{zhu2024improving} employs an entropy-based confidence metric to optimize candidate selection. While these sampling methods effectively reduce degeneration, they risk inconsistency, compromising overall text quality \cite{basubadsampling}.

\paragraph{Contrastive Strategies} typically aim at balancing coherence and diversity in text generation tasks. Contrastive Decoding \citep[CD,][]{cd} leverages the differential between expert and amateur language models to select tokens that maximize their log-likelihood difference. CS evaluates top-$k$ tokens using a combination of model confidence and degeneration penalty to enhance output diversity. ACS builds upon CS by eliminating hyperparameters, addressing the unexplored impact of hyperparameter selection on generation quality \cite{garces-arias-etal-2025-decoding}, at the cost of additional computational overhead. Moreover, its focus on immediate uncertainty may overlook temporal fluctuations that could benefit from more robust generation strategies \cite{chen2024decodinggameminimaxoptimality}. These limitations motivate GUARD, a novel decoding strategy that offers an efficient, self-adaptive decoding guided by uncertainty.

\section{Methodology}
\label{sec:methodology}

Figure \ref{fig:figure_two} illustrates GUARD, which (a) leverages global entropy $H_{\text{glob, $t$}}$ (Eq. \eqref{eq:hglob}) to capture model uncertainty fluctuations throughout the decoding process and (b) introduces Glocal uncertainty (denominator in Eq. \eqref{eq:kt}). It thus considers both long- and short-term uncertainty fluctuations to enable adaptive control over $k_t$ and $\alpha_t$ during (A)CS for the top-$k_t$ most semantically accurate tokens. The method incorporates a simple yet effective token count-based penalty for repeated tokens.

\begin{figure*}[!ht] 
    \centering
    \includegraphics[width=\textwidth]{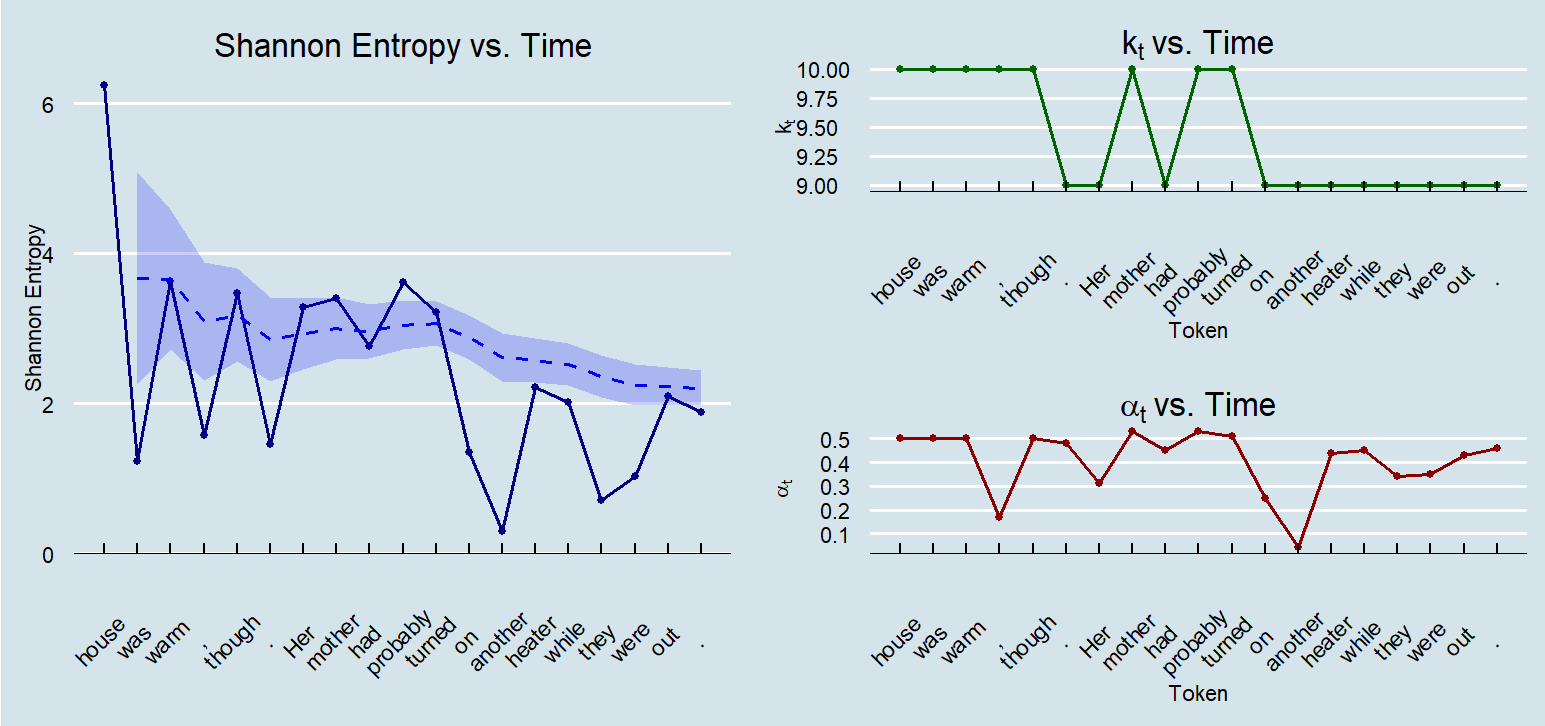}
    \begin{minipage}{1\textwidth}
        \small
        \raggedright 
        \textbf{Prompt:} \textit{A shiver ran through her and she walked back inside. There was a chill in the
air and she was only wearing a t-shirt and jeans. The}
  \\        
       \textbf{Generated story:} \textit{house was warm, though. Her mother had probably turned on another heater while they were out.}
    \end{minipage}
    \caption{\textbf{Left:} Local (solid) and global (dashed/interval) Shannon entropy over the time steps of the generation. \textbf{Right:} Strong changes are smoothed to provide a robust entropy estimation before computing $k$ and $\alpha$ over time, leading to increased stability.} 
    \label{fig:decoding_behavior} 
\end{figure*}

\subsection{Global Entropy and Statistical Properties}\label{subsec:global-entropy}

In the related ACS strategy, the Shannon entropy $H(X)_{t}=-\sum_{x \in \mathcal{V}} p(x\mid\boldsymbol{x}_{<t}) \ln p(x\mid\boldsymbol{x}_{<t})$ is used to dynamically adjust hyperparameters at the time step $t$. As is customary, $X$ denotes a discrete random variable with $\mathcal{V}$ referring to the tokens; $t \in \{1, \dots, T\}$ denotes a single generation step, limited to the maximum sequence length (e.g., $T = 256$).

However, the Shannon entropy $H(X)_{t}$ only measures \textit{local} uncertainty, i.e., the entropy at each generation time step $t \in \{1, \dots, T\}$. As illustrated in Figure~\ref{fig:decoding_behavior} (left, solid line), local entropy measurements exhibit marked volatility. Addressing \textbf{RQ2}, we investigate whether these abrupt \say{spikes} can be effectively attenuated by a moving average of instantaneous entropy values, which we define as \textbf{global entropy} (dashed line, left part of Figure~\ref{fig:decoding_behavior}).
We define global entropy $H_{\text{glob, $t$}}$ at time step $t$ in Eq. \eqref{eq:hglob} as the weighted average of the Shannon entropies (measuring local uncertainty) throughout the generation process:
\begin{equation}
    H_{\text{glob}, t} = \frac{\sum_{i=1}^{t} (\lambda^{t-i} \cdot H(X)_{i})}{\sum_{i=1}^{t} \lambda^{t-i}}
\label{eq:hglob}
\end{equation}

The parameter $\lambda \in (0,1]$ denotes a weighting coefficient. We employ \( \lambda^{t-i} \) to reflect the temporal evolution of model uncertainty over time (i.e., over sequential generation steps), which can dynamically adjust the contribution of each generation step to \( H_{\text{glob}, t} \).  

In the following, we will provide Proposition~1 to prove that $H_{\text{glob}, t}$ is an unbiased estimator of the instantaneous entropy $H(X)_{t}$, irrespective of the concrete choice of $\lambda \in (0,1]$. In Proposition~2, we further prove that $H_{\text{glob}, t}$ is a consistent estimator of $H(X)_{t}$ under reasonable assumptions. This tells us that $H_{\text{glob}, t}$ converges in probability, i.e., as $t \rightarrow \infty$, to the instantaneous entropy. Thus, it is not only unbiased but also has diminishing variance when the generation is sufficiently long. Detailed proofs of both results can be found in Appendix \ref{app:A}.

\paragraph{Proposition 1 (Unbiasedness).}
Assume that the process \(\{H(X)_t\}_{t\geq 1}\) is stationary. Then, 
\begin{align}
    \mathbb E\bigl[H_{\text{glob}, t}\bigr] = \mathbb E[ H(X)_t]
    \label{ex}
\end{align}
for any $\lambda \in [0,1)$.

\paragraph{Sketch of Proof.}
The result directly follows from the expectation's linearity and \(\{H(X)_t\}_{t\ge 1}\) being stationary. Linearity gives
\begin{equation}
    \mathbb E\bigl[H_{\text{glob}, t}\bigr] = \sum_{i=1}^{t} \frac{\lambda^{t-i}}{\sum_{s=1}^{t} \lambda^{t-s}}\, \mathbb E\bigl[H(X)_i\bigr]
\label{et}
\end{equation}
Stationarity implies
\begin{equation}
     \forall t \in \{1, \dots, T\}: \;  \mathbb E\bigl[H(X)_t\bigr] = H_0,
\end{equation}
from which the claim obviously follows. $\hfill \square$

\paragraph{Proposition 2 (Consistency).}
Assume the process \(\{H(X)_{t}\}_{t\ge 1}\) is stationary, ergodic and bounded. 
Further assume that the covariances \(\operatorname{Cov}\bigl(H(X)_{t},H(X)_{s}\bigr)\) decay sufficiently fast as \(|t-s|\to\infty\). Then, as $t \rightarrow \infty$,
\begin{equation}
    H_{\text{glob}} \stackrel{\mathbb P}{\longrightarrow} \mathbb E[H(X)_{t}]
    \label{ex2}
\end{equation}

\paragraph{Sketch of Proof.}
We have that all \(H(X)_{t}\) are bounded and their covariances decay sufficiently fast as \(|t-s|\) increases. Moreover, the weights form a convex combination. Under standard mixing conditions, we show that
\[
\operatorname{Var}\bigl(H_{\text{glob}}\bigr) = \mathcal O\Bigl(\frac{1}{t}\Bigr),
\]
where $\mathcal O$ is the Landau (\say{big O}) notation.
Then, by Chebyshev's inequality, for any \(\epsilon>0\) we have
\[
\mathbb P\Bigl(\bigl|H_{\text{glob}} - \mathbb E[H_{\text{glob}}]\bigr| > \epsilon\Bigr) \le \frac{\operatorname{Var}\bigl(H_{\text{glob}}\bigr)}{\epsilon^2} \to 0 
\]
as $t \to \infty$, from which the claim follows. \hfill $\square$.

\paragraph{Brief Summary.} In Proposition~1 and Proposition~2, we prove that $H_{\text{glob}, t}$ is not only unbiased, but also has diminishing variance when the generation is long enough. Such statistical properties support that we can reasonably estimate the instantaneous entropy with $H_{\text{glob}, t}$. This answers \textbf{RQ2} -- the intended smoothing of the uncertainty estimation does not come at the price of losing statistical validity. However, note that both results concern our internal estimators of the entropy, not the inferential properties of the final language model. 

\paragraph{Interpretation.} Our method employs \( \lambda^{(t-i)} \) to reflect the temporal evolution of model uncertainty. Specifically, as the time step \( i \) approaches the current moment \( t \), its contribution to \( H_{\text{glob}} \) increases. Conversely, time steps distant from \( t \) have diminishing effects on \( H_{\text{glob}} \). The \( \lambda^{(t-i)} \) factor implements a temporal decay mechanism, dynamically adjusting each time step's contribution to \( H_{\text{glob}} \). We later set $\lambda = 0.95$, but recall that our method's unbiasedness and consistency hold for any $\lambda \in (0,1]$.\footnote{We experimentally verified that performance is relatively insensitive to the chosen value of $\lambda$ (cf. Tab. \ref{tab:lambda_selection}, Appendix \ref{app:B}) and thus set $\lambda=0.95$ as an intermediate value.} Therefore, our Global entropy measure can effectively estimate local uncertainties in a statistically unbiased and consistent way.

\subsection{Glocal Uncertainty and Candidate Token Set}
\label{subsec:efficiency}
To determine \(k_t\), that is, the selection of candidate tokens in CS, we introduce \textit{Glocal} uncertainty. It simultaneously considers both the \textit{global} and \textit{local} uncertainty fluctuations in the token generation steps, denoted as \(\delta_{glob}\) and \(\delta_{loc}\), respectively. Specifically, we present Eq. \eqref{eq:kt} to automatically derive \(k_t\) for the Top-\( k_t \) candidate tokens. 
\begin{equation}
\scriptstyle
    k_t = 10 \cdot (1 - \frac{1}{\exp\left( \lambda_k \cdot \delta_{\text{loc}} + (1 - \lambda_k) \cdot \delta_{\text{glob}} \right) + 1}) + 5
    \label{eq:kt}
\end{equation}
\begin{equation}
\scriptstyle
    \delta_{\text{loc}} = q \cdot \operatorname{arctanh} \left( \frac{H(X)_t - \operatorname{med}\left( H(X)_{t-w:t-1} \right)}{\ln |\mathcal V|} \right)
\label{eq:deltaloc}
\end{equation}
\begin{equation}
\scriptstyle
    \delta_{\text{glob}} = q\cdot \operatorname{arctanh}\left( 
    \frac{
        \operatorname{med}\left( H(X)_{t-w:t-1} \right) 
        -\operatorname{med}\left( H_{\text{glob}, t} \right)
    }
    {\ln|\mathcal V|}
    \right)
\label{eq:deltaglob}
\end{equation}
\begin{equation}
    \lambda_k = \frac{|\delta_{\text{loc}}|}{|\delta_{\text{loc}}| + |\delta_{\text{glob}}| + \epsilon}
    \label{eq:lanbda}
\end{equation}

where the constants 10 and 5 for the parameter $k_t$ define an effective range for exploration \citep{estebanACS}. \( \delta_{\text{loc}} \) measures the model's instantaneous uncertainty fluctuation compared to recent steps, i.e.,  the instantaneous uncertainty at the current $t^{th}$ step $H(X)_t$  versus the median uncertainty in the recent time window \( w \),\footnote{To select the optimal value for \( w \), we conducted extensive experiments, which are presented in Appendix \ref{app:B} (Tables \ref{tab:ablation_selectw} -- \ref{tab:ablation_selectw_gpt2}). We observe that \( w \) does not notably influence text quality. Ultimately, we set $w = 7$, keeping it constant for our experiments.} denoted as $\operatorname{\textit{med}}\left( H(X)_{t-w:t-1} \right)$. \( \delta_{\text{glob}} \) assesses the overall variation between the median of the short-term entropy $\operatorname{\textit{med}}\left( H(X)_{t-w:t-1} \right)$ and the median of the long-term uncertainty trend up to the current $t^{th}$ step $\operatorname{\textit{med}}\left(H_{\text{glob}, t}\right)$.\footnote{The choice of \textit{median} and \textit{arctanh} was experimentally validated. Our experiments compared different pairings, including \textit{winsorized mean}, \textit{mean}, \textit{median}, \textit{logarithmic map}, and \textit{arctanh}, with detailed results shown in Table \ref{tab:median_selection}, Appendix \ref{app:B}.} $\lambda_k$ refers to the dynamic adjustment of local and global entropy during the generation process. \( q \) is an adaptive temperature parameter influencing the dynamic adjustment magnitude of \( k_t \) and \( \alpha_t \). It adaptively adjusts by integrating changes in both current and global uncertainties:
\begin{equation}
    q = 1.0 + r_{\text{change}} + r_{\text{difference}}
    \label{eq:q}
\end{equation}

\( r_{\text{change}} \) measures the rate of uncertainty change at $t$:
\begin{equation}
    r_{\text{change}} = \frac{ | H(X)_{t} - H(X)_{t-1} | }{ H(X)_{t-1} + \epsilon }
    \label{eq:rchange}
\end{equation}

\( r_{\text{difference}} \) assesses the difference between local uncertainty within the current time window and the overall trend:
\begin{equation}
    r_{\text{difference}} = \frac{ | \text{med}( H(X)_{t-w:t-1} ) - \text{med}( H_{\text{\text{glob},t}} ) | }{ \text{med}( H(X)_{t-w:t-1} ) + \epsilon }
    \label{eq:rdiff}
\end{equation}

When \( r_{\text{change}} \) or \( r_{\text{difference}} \) is large, \( q \) increases, causing \( \delta_{\text{loc}} \) and \( \delta_{\text{glob}} \) to be amplified. \( \lambda_k \) and \( \lambda_\alpha \) dynamically allocate weights based on their ratio to adjust \( k_t \) and \( \alpha_t \), which are part of the token count penalty. When uncertainty is high, \( k_t \) increases, potentially introducing repetitive candidate words, but also \( \alpha_t \) increases, amplifying the degradation penalty and counteracting the effect of a higher \( k_t \). Conversely, a decrease of \( k_t \) and \( \alpha_t \) results in outputs that favor high-confidence candidate words.

\paragraph{Rationale of Glocal Uncertainty.} In Eq. \eqref{eq:kt}, the denominator $C_{\text{sum}} = \exp( \lambda_k \cdot \delta_{\text{loc}} + (1 - \lambda_k) \cdot \delta_{\text{glob}})$ can be regarded as an alternative to perplexity \cite{ppl}, since the latter is the exponential of the total uncertainty in the prediction up to the current $t^{th}$ step, where Glocal uncertainty calculates the variations between the instantaneous entropy at the $t^{th}$ step with respect to short-term median entropy, as well as the short- vs. longer-term entropy variations. Similar to perplexity, a lower $C_{\text{sum}}$ value is preferred. When pronounced uncertainty variations occur, higher perplexity is indicated, suggesting we should select a larger \(k_t\) value to expand the candidate token set, accommodating model instability. Conversely, when variations are minimal, a smaller \(k_t\) value suffices, as the model demonstrates stability, allowing us to evaluate fewer candidate tokens.

Once \(k_t\) is determined, we select the adequate Top-\( k_t \) candidate tokens $\mathcal V^{(k_t)}$, then design a token count penalty strategy to increase the diversity of the generated texts by reducing repetitiveness. Rather than employing a cosine-similarity-based penalty as in (A)CS, GUARD (as defined in Eq. \eqref{eq:acs}) circumvents the over-penalization of semantic similarity, leads to a more effective reduction of repetitions (and hence increased diversity), and accelerates the generation speed:
\begin{align}
\scriptstyle
    P(x_{next}) = \underset{v \in \mathcal V^{(k_t)}}{\arg \max} \Bigg\{ {p_\theta(v \mid \boldsymbol{x}_{<k_t})} \times
    \underbrace{\alpha_t ^ {token \ counts(v)}}_{\text{degeneration penalty}} \Bigg\}
    \label{eq:acs}
\end{align}

where \( \alpha_t \) is the penalty coefficient and \( \text{token\ counts}(v) \) denotes the frequency of the candidate \( v \) in the previously generated sequence. To dynamically determine $\alpha_t$, we reuse the core of Eq. \eqref{eq:kt}, replacing $|\mathcal V|$ with $k_t$ in Eq. \eqref{eq:deltaloc} and \eqref{eq:deltaglob}:
\begin{equation}
    \alpha_t = 1 - \frac{1}{\exp\left( \lambda_k \cdot \delta_{\text{loc}} + (1 - \lambda_k) \cdot \delta_{\text{glob}} \right) + 1}
    \label{eq:alphat}
\end{equation}

When $\mathcal V^{(k_t)}$ is small, each repetitive token in $\mathcal V^{(k_t)}$ will receive a very light penalty, and vice versa. Appendix \ref{app:C} illustrates the entire algorithm as pseudo-code to facilitate understanding of its flow. In Figure \ref{fig:decoding_behavior}, we demonstrate that fluctuations can be smoothed to ensure a robust entropy estimation (left) when using our technique in determining $k_t$ and \( \alpha_t \) over time (right). Our experiments show that the token count-based penalty strategy can substantially reduce repetitions, therefore increasing the diversity of the text outputs while enhancing text generation speed compared to using cosine similarity. 

\paragraph{Self-Adaptive Design of GUARD.} GUARD achieves self-adaptivity for all quality-determining hyperparameters in CS methods; while it exhibits hyperparameters $\lambda$ and $w$, extensive testing (Appendix \ref{app:B}) confirms that specific choices have negligible impact on generation quality, unlike other decoding methods \citep{garces-arias-etal-2025-decoding}. 

\section{Experiments}
\label{sec:exp_setup}

\subsection{Experimental Setup}
This section will introduce the metrics and datasets used in our experiments, as well as the baseline decoding strategies we use for reference and the LLMs we employ.

\begin{table*}[!ht]
\centering
\resizebox{1\textwidth}{!}{%
\begin{tabular}{
      l
      *{4}{S[table-format=2.2]}
      *{4}{S[table-format=2.2]}
      *{4}{S[table-format=2.2]}
    }
      \toprule
      \multirow{2}{*}{Strategy}
        & \multicolumn{4}{c}{Wikitext (\textit{n = 1314})}
        & \multicolumn{4}{c}{Wikinews (\textit{n = 2000})}
        & \multicolumn{4}{c}{BookCorpus (\textit{n = 1947})}\\
      \cmidrule(lr){2-5} \cmidrule(lr){6-9} \cmidrule(lr){10-13}
        & {Div.} & {MAUVE} & {Coh.} & {BERT-Sc.}
        & {Div.} & {MAUVE} & {Coh.} & {BERT-Sc.}
        & {Div.} & {MAUVE} & {Coh.} & {BERT-Sc.} \\
      \midrule
      \rowcolors{1}{gray!10}{white}

      Greedy
        & 16.97 & 48.28 & -0.78 & 81.45
        & 28.13 & 74.91 & -0.90 & 82.51
        &  4.07 & 17.99 & -0.65 & 79.92 \\

      Beam ($5$)
        & 14.42 & 26.39 & -0.77 & 81.20
        & 24.94 & 54.31 & -0.87 & 83.23
        &  5.39 & 18.69 & -0.56 & 80.03 \\

      Temp. ($0.9$)
        & 84.81 & 92.85 & \colorbox{Best}{\textbf{-2.43}} & 81.86
        & 92.71 & \colorbox{Best}{\textbf{96.29}} & -2.41 & 83.16
        & 89.58 & 85.54 & -2.33  & 81.78 \\

      Top-\(k\) (\(50\))
        & 82.69 & 88.97 & -1.98 & 80.64
        & 92.13 & 94.52 & -2.01 & 83.17
        & 91.54 & 93.24 & -2.52 & 81.29 \\

      Top-\(p\) (\(0.95\))
        & 82.47 & 86.37 & -2.11 & 81.50
        & 91.47 & 96.09 & -2.15 & 82.84
        & 88.92 & \colorbox{Best}{\textbf{94.81}} & -2.29 & 81.31 \\

      Typical
        & 69.54 & 81.57 & -1.78 & 81.34
        & 82.31 & 95.66 & -1.83 & \colorbox{Best}{\textbf{83.86}}
        & 96.29 & 88.58 & -3.68 & 80.06 \\

      CD
        & 72.70 & 79.76 & -2.61 & 81.72
        & 75.28 & 76.20 & -2.36 & 82.57
        & 76.00 & 72.10 & \colorbox{Best}{\textbf{-2.53}} & 81.51 \\

      AD
        & 87.19 &  \colorbox{Best}{\textbf{92.20}} & -2.51 & 81.83
        & 91.49 & 94.38 & -2.35 & 82.52
        & 90.70 & 94.49 & -2.49 & 81.82 \\

      CS (\(5,0.6\))
        & 72.08 & 80.52 & -1.43 & 81.53
        & 87.47 & 91.47 & -1.31 & 83.01
        & 68.26 & 87.92 & -1.24 & 81.20 \\

      CS (\(10,0.6\))
        & 78.68 & 73.50 & -1.69 & 80.57
        & 91.38 & 90.49 & -1.57 & 83.09
        & 82.60 & 75.53 & -1.57 & 81.22 \\

      ACS (\(q=1\))
        & 81.57 & 78.82 & -1.95 & 80.00
        & \colorbox{Best}{\textbf{92.79}} & 91.23 & -1.72 & 81.50
        & 87.86 & 78.32 & -1.52 & 79.79 \\

      \textit{GUARD}
        & \colorbox{Best}{\textbf{92.86}} & 90.82 & -2.61 & \colorbox{Best}{\textbf{81.90}}
        & 95.20 & 93.60 & \colorbox{Best}{\textbf{-2.38}} & 83.16 
        & \colorbox{Best}{\textbf{96.18}} & 92.59 & \colorbox{Best}{\textbf{-2.53}} & \colorbox{Best}{\textbf{81.83}} \\

      \midrule

      Human
        & 93.84 & 100.00 & -2.43 & 100.00
        & 93.50 & 100.00 & -2.86 & 100.00
        & 94.98 & 100.00 & -2.99 & 100.00 \\
      \bottomrule
    \end{tabular}%
}
\caption{Averaged automatic evaluation results for Qwen2.5-7B. Hyperparameters for competing strategies chosen based on experimental evaluation (Appendix \ref{app:G}, Tables \ref{tab:different_beam_width}, \ref{tab:different_topk}, \ref{tab:different_topp}) or based on the original paper (CS, ACS). Scores closest to human highlighted in {\colorbox{Best}{\textbf{bold}}}.}
\label{tab:ablation_resultglacs}
\end{table*}

\begin{table}[!ht]
\centering
\resizebox{0.5\textwidth}{!}{
\begin{tabular}{ccccc}
\toprule
\multirow{2}{*}{Dataset} & \multicolumn{4}{c}{Human Evaluation}  \\
\cmidrule(lr){2-5} 
(\textit{n = 200} & \multicolumn{2}{c}{Sem. Coh.* (\%)↑} & \multicolumn{2}{c}{Fluency**  (\%)↑}  \\
 each)& ACS & GUARD & ACS & GUARD  \\
\midrule
Wikitext & 37.50 & {\colorbox{Best}{\textbf{62.50}}} & 43.75 & \colorbox{Best}{\textbf{56.25}}  \\
Wikinews & 40.00 & {\colorbox{Best}{\textbf{60.00}}} & 45.25 & {\colorbox{Best}{\textbf{54.75}}}  \\
BookCorpus & 48.75 & {\colorbox{Best}{\textbf{51.25}}} & {\colorbox{Best}{\textbf{52.50}}} & 47.50   \\
\midrule
All & 42.00 & \colorbox{Best}{\textbf{58.00}} & 47.00 & {\colorbox{Best}{\textbf{53.00}}}  \\
\bottomrule
\end{tabular}
}
\caption{\textbf{Human evaluation:} Share of human evaluators favoring a strategy w.r.t. perceived semantic coherence and fluency. P-values for the exact binomial test: *$p_{\text{Sem. Coh.}}=2.875e-06$ , **$p_{\text{Fluency}}=0.01819$. Best results are highlighted in \colorbox{Best}{\textbf{bold}}.}
\label{tab:humaneval_resultglacs}
\end{table}

\begin{table*}[!ht]
\centering
\resizebox{1\textwidth}{!}{
\begin{tabular}{cccc|ccc|ccc}
\toprule
 Metric & GUARD wins & ACS wins & Tie & GUARD wins & Top-$k$ wins & Tie & GUARD wins & Top-$p$ wins & Tie \\
\midrule
Overall  & \colorbox{Best}{\textbf{34.3}} & 30.6 & 35.1 & \colorbox{Best}{\textbf{65.8}} & 17.5 & 16.7 & \colorbox{Best}{\textbf{49.2}} & 34.2 & 16.7 \\
Fluency & \colorbox{Best}{\textbf{40.7}} & 35.3 & 24.0 & \colorbox{Best}{\textbf{76.7}} & 23.3 & 0.0 & \colorbox{Best}{\textbf{56.7}} & 43.3 & 0.0 \\
Sem. Coh. & \colorbox{Best}{\textbf{40.7}} & 34.0 & 25.3 & \colorbox{Best}{\textbf{80.0}} & 20.0 & 0.0 & \colorbox{Best}{\textbf{58.3}} & 41.7 & 0.0 \\
Factuality & 4.0 & \colorbox{Best}{\textbf{10.0}} & 86.0 & 0.0 & 0.0 & \colorbox{Best}{\textbf{100.0}} & 0.0 & 0.0 & \colorbox{Best}{\textbf{100.0}} \\
Informativeness & \colorbox{Best}{\textbf{40.0}} & 35.3 & 24.7 & \colorbox{Best}{\textbf{78.3}} & 21.7 & 0.0 & \colorbox{Best}{\textbf{60.0}} & 40.0 & 0.0 \\
Interestingness & \colorbox{Best}{\textbf{40.0}} & 34.7 & 25.3 & \colorbox{Best}{\textbf{80.0}} & 20.0 & 0.0 & \colorbox{Best}{\textbf{60.0}} & 40.0 & 0.0 \\
Story Development & \colorbox{Best}{\textbf{40.7}} & 34.0 & 25.3 & \colorbox{Best}{\textbf{80.0}} & 20.0 & 0.0 & \colorbox{Best}{\textbf{60.0}} & 40.0 & 0.0 \\
\bottomrule
\end{tabular}

}
\caption{Win shares (\%) of GUARD vs. ACS/Top-$k$/Top-$p$ for LLM evaluation for $n=150$ continuations. Best results are highlighted in \colorbox{Best}{\textbf{bold}}.}
\label{tab:llm_judgements}
\end{table*}

\paragraph{LLMs.} We employ six LLMs: Qwen2.5-7B \cite{qwen2.5}, Deepseek-llm-7B-base \cite{deepseekai2024deepseekv3technicalreport}, Mistral-v0.3 \cite{jiang2023mistral7b}, Llama-3 \cite{dubey2024llama3herdmodels}, Llama-2 \cite{llama2}, Gemma-7B \cite{gemma}.

\paragraph{Evaluation Metrics.} We employ the following five commonly used metrics to automatically assess generation quality: Diversity \cite{su2022contrastive}, MAUVE \cite{mauve}, Coherence \cite{su2022contrastive}, BERTScore \cite{bert-score}. Formulas are provided in Appendix \ref{app:D}. 

\paragraph{Human Evaluation.} To evaluate the quality of the generated text, human evaluators assessed two key aspects: semantic coherence and fluency. Four native English speakers evaluated 600 pairs of competing text continuations, which were uniformly distributed across the three datasets. We measured inter-rater agreement using Fleiss' Kappa and performed exact binomial tests to determine statistically significant differences between ACS and GUARD. Further details on the evaluation are provided in Appendix \ref{app:E}.

\paragraph{LLM Evaluation.} We employ LLM-as-a-judge (judge: GPT-4) to systematically evaluate the quality of the generated text on $n=150$ continuations, uniformly distributed across the three datasets. This method aims to approximate the evaluation process of human experts on a large scale. For this, we measured six core metrics: fluency, coherence, factuality, informativeness, interestingness, and story development. The prompt is provided in Appendix \ref{app:F}.

\paragraph{Datasets and Baselines.} We focus on three datasets for open-ended text generation from different domains: Wikinews (\textit{n = 2000}), Wikitext \citep[\textit{n = 1314; }][]{wikitext1314}, and BookCorpus \citep[\textit{n = 1947; }][]{book1974}. We compare GUARD to SOTA LLM decoding methods from three categories: Deterministic (greedy search, beam search with $B \in \{3, 5, 10, 15, 20, 50\}$), stochastic (temperature sampling with $\tau = 0.9$, top-$k$ sampling with $k \in \{3, 5, 10, 15, 20, 50\}$, top-$p$ sampling with $p \in \{0.60, 0.70, 0.80, 0.90, 0.95\}$, typical sampling with $\tau = 0.2$, adaptive decoding), and contrastive approaches (CD, CS with $k \in \{5, 10\}$, $\alpha = 0.6$, and ACS with $q=1$). Across all experiments, we used a maximum token length of 256 tokens for the text generations.

\paragraph{Hardware.} We used NVIDIA 4090 (24GB memory) and NVIDIA H100 (80GB memory).

\subsection{Results}
\label{sec:results}

\paragraph{Automatic evaluation results.} The automatic evaluations of different methods, based on diversity, MAUVE, and coherence, are presented in Table \ref{tab:ablation_resultglacs}. We observe that, in general, GUARD outperforms all other competing strategies in both diversity and coherence. In terms of MAUVE, it is roughly on par with the best competing methods on each dataset. More importantly, it achieves results closest to human references without relying on specific hyperparameter choices. We observe that GUARD generally performs well according to BERTScore, which, however, is not very discriminative. Furthermore, we further validate the robustness of our method for other model architectures (cf. Appendix \ref{app:G}, Tables \ref{tab:model_size_results} -- \ref{tab:model_size_bertscore_results}). Notably, GUARD is well-suited for decreasing the repetitive use of single tokens, as indicated by the high diversity values across all three benchmark datasets. While one might suspect that this improvement comes at the cost of coherence, our experiments show that GUARD also consistently exhibits coherence values that are very close to those of the human-written completions.

\begin{table*}[!htb]
\centering
\resizebox{1\textwidth}{!}{
\begin{tabular}{llccccccc}
\toprule
\multirow{2}{*}{Metric} & \multirow{2}{*}{Dataset} & \multicolumn{3}{c}{Human Evaluation} & \multicolumn{3}{c}{LLM Evaluation} & \multirow{2}{*}{Agree?} \\
\cmidrule(lr){3-5} \cmidrule(lr){6-8}
 & ($n=50$ each) & GUARD (\%) & ACS (\%) & Ties (\%) & GUARD (\%) & ACS (\%) & Ties (\%) &   \\
\midrule
\multirow{4}{*}{Sem. Coh.} 
 & Wikitext & {\colorbox{Best}{\textbf{44.0}}} & 19.0 & 37.0 & {\colorbox{Best}{\textbf{42.0}}} & 32.0 & 26.0 & $\checkmark$ \\
 & Wikinews & {\colorbox{Best}{\textbf{43.5}}} & 23.5 & 33.0 & {\colorbox{Best}{\textbf{40.0}}} & 34.0 & 26.0 & $\checkmark$\\
 & BookCorpus  & {\colorbox{Best}{\textbf{38.0}}} & 35.5 & 26.5 & {\colorbox{Best}{\textbf{40.0}}} & 36.0 & 24.0 & $\checkmark$ \\
 \cmidrule(lr){2-9}
 & Overall ($n = 150$) & {\colorbox{Best}{\textbf{41.8}}} & 26.0 & 32.2 & {\colorbox{Best}{\textbf{40.7}}} & 34.0 & 25.3 & $\checkmark$ \\
 \midrule\midrule
 \multirow{4}{*}{Fluency} 
 & Wikitext & {\colorbox{Best}{\textbf{23.0}}} & 10.5 & 66.5 & {\colorbox{Best}{\textbf{42.0}}} & 32.0 & 26.0 & $\checkmark$ \\
 & Wikinews & {\colorbox{Best}{\textbf{21.5}}} & 12.0 & 66.5 & {\colorbox{Best}{\textbf{40.0}}} & {\colorbox{Best}{\textbf{40.0}}} & 20.0 & $\tikzxmark$ \\
 & BookCorpus  & 13.0 & {\colorbox{Best}{\textbf{18.0}}} & 69.0 & {\colorbox{Best}{\textbf{40.0}}} & 34.0 & 26.0 & $\tikzxmark$ \\
 \cmidrule(lr){2-9}
 & Overall ($n = 150$) & {\colorbox{Best}{\textbf{19.2}}} & 13.5 & 67.3 & {\colorbox{Best}{\textbf{40.7}}} & 35.3 & 24.0 & $\checkmark$\\
\bottomrule
\end{tabular}

}
\caption{Win shares (Human and LLM evaluation) w.r.t. coherence and fluency of GUARD and ACS. Best results are highlighted in \colorbox{Best}{\textbf{bold}}.}
\label{tab:llm_with_human}
\end{table*}

\paragraph{Human evaluation.} We only compare ACS and GUARD for the sake of practicability and since both show strong performance (cf. Table \ref{tab:ablation_resultglacs}. The results of this study are summarized in Table~\ref{tab:humaneval_resultglacs}: Overall, GUARD demonstrates superior performance compared to ACS in terms of both semantic coherence and fluency, with the sole exception of fluency on the BookCorpus dataset. Specifically, human preferences indicate that GUARD outperforms ACS on average by approximately 6\% in semantic coherence and 16\% in fluency across all datasets. Both differences are statistically significant at the $\alpha = 0.05$ level. The inter-rater reliability (Fleiss' $\kappa$ = 0.41) shows moderate agreement among evaluators. In Appendix \ref{app:H}, we include a small case study for the interested reader that further substantiates these findings. Overall, the human evaluation comprehensively addresses \textbf{RQ1}, confirming that both automatic metrics and human judgments favor our approach over alternative methods.

\paragraph{LLM evaluation.} We include LLM evaluation as an approximation of human judgments, comparing GUARD to its natural baseline ACS, as well as Top-$k$ and Top-$p$. The evaluation focuses on the six dimensions described in the experimental setup. The results (cf. Table \ref{tab:llm_judgements}) demonstrate that GUARD outperforms all three competitors in five out of six metrics (except factuality), where ACS has a slight edge, and comparison to the other two strategies results in a tie. When comparing human to LLM judgments for coherence and fluency (cf. Table \ref{tab:llm_with_human}), results (a) demonstrate strong agreement and (b) show that human evaluators indicate an even stronger preference for GUARD's coherence compared to LLM-judges. 

\begin{table}[!htb]
\centering
\resizebox{0.5\textwidth}{!}{
\begin{tabular}{cccc}
\toprule
\textbf{Method} & \textbf{sec/story} & \textbf{\#tokens/sec} & \textbf{\#tokens/story} \\
\midrule
CS (10, 0.6) & 11.6 & 22.0 & 249.7\\
ACS ($q$ = 1) & 15.7 & 16.3 & 255.7 \\
ACS ($q$ = 2) & 15.9 & 16.1 & 255.8\\
ACS ($q$ = 8) & 16.3 & 15.3 & 255.8 \\
\textit{GUARD} & {\cellcolor{Best}\textbf{4.42}} & {\cellcolor{Best}\textbf{28.3}} & \cellcolor{Best}\textbf{255.9}\\
\bottomrule
\end{tabular}
}
\caption{Average generation speed ($n = 30$) of GUARD with CS and ACS ($q \in \{1, 2, 8\}$). Best results in \colorbox{Best}{\textbf{bold}}.}
\label{tab:speed_comparison}
\end{table}

\paragraph{Generation speed.} In Table \ref{tab:speed_comparison}, we compare GUARD to CS and ACS in terms of the average generation speed. Based on 30 generation samples, all three methods produce generations of comparable lengths. CS spends 11.6 seconds per story, ACS takes an even longer, from 15.7 seconds to 16.3 seconds (across different temperature settings), while GUARD only needs 4.42 seconds per story. We measure the average number of tokens decoded per second, and observe that CS decodes on average 21.98 tokens/second, ACS between 15.35 and 16.29, and GUARD decodes 28.3 tokens/second on average. 
This answers \textbf{RQ3}, since the generation speed of CS-based approaches can be substantially increased using GUARD. 

\section{Discussion}

\paragraph{Stationarity Assumption.} The strict stationarity assumption might be a strong idealization in real, dynamic text generation processes. However, the main purpose of introducing the stationarity assumption here is to provide a theoretical guarantee of unbiasedness for our $H_{glob,t}$ estimator (Proposition~1). This theoretical result aims to show that, under ideal conditions, our weighted average design does not introduce systematic bias. It is worth noting that the core advantage of our work does not entirely rely on this strict assumption. The key role of our design (i.e., the exponentially weighted moving average in Eq. \eqref{eq:hglob}) in practice is to smooth out drastic fluctuations in entropy. The presence of the decay factor $\lambda$ makes it more sensitive to recent entropy values, while gradually \say{forgetting} older ones. This mechanism allows $H_{glob,t}$ to dynamically adapt to local changes in the entropy process, providing a locally stationary and robust long-term trend estimate, even if the overall process is non-stationary. As shown in Figure \ref{fig:decoding_behavior}, $H_{glob,t}$ effectively smooths out local entropy, illustrating that our method remains effective even in the presence of entropy fluctuations (non-stationary conditions).

\paragraph{Self-Adaptiveness.} While we rightfully advertise our method's ability to dynamically adjust core decoding hyperparameters ($k$, $\alpha$, and $q$) without manual tuning, GUARD still relies on $\lambda$ and $w$. Nevertheless, our experiments suggest that neither of them substantially influences performance, and can be kept at their default values. One point to note for practical use is that we only validated this for generations of length up to 256 tokens.

\section{Conclusion}
\label{sec:conclusion}

We introduce GUARD, a self-adaptive decoding strategy that adaptively balances text coherence and diversity by responding to fluctuations in long-term and short-term uncertainty. By integrating a token frequency penalty into CS, GUARD reduces repetitive outputs and computational overhead while maintaining context fidelity. Theoretical analysis confirms that our estimation of global entropy preserves key statistical properties, such as unbiasedness and consistency. Comprehensive experiments across multiple open-source models, datasets, metrics, human and LLM evaluations demonstrate improvements in coherence, fluency, diversity, efficiency, informativeness, interestingness, and story development over established decoding methods. Overall, GUARD offers a robust approach for improving open-ended text generation. Future work will extend our method to additional text-generation tasks. We believe that investigating its integration with supervised fine-tuned models and exploring its performance in multilingual and low-resource settings might yield valuable insights.

\section*{Limitations}
\label{sec:limitations}

While our proposed method demonstrates improvements in open-ended text generation quality, several limitations warrant acknowledgment:
\paragraph{1.} Our evaluation focuses exclusively on open-ended text generation tasks. The transferability of our approach to other NLP applications, such as summarization, machine translation, and classification, remains to be investigated.
\paragraph{2.} The empirical evaluation is confined to English-language datasets, leaving questions about cross-lingual generalizability, particularly for low-resource languages, unanswered.
\paragraph{3.} While we demonstrate effectiveness across various open-source models, we have not evaluated our method on proprietary language models, which may exhibit different behaviors.
\paragraph{4.} Our experiments are limited to base models. The impact of our approach on supervised fine-tuned models represents an important direction for future research.
\paragraph{5.} Our setting was limited by a maximum length of 256 tokens, and the behaviour in long-text generation scenarios is yet to be explored.

\section*{Ethics Statement}

We affirm that our research adheres to the \href{https://www.aclweb.org/portal/content/acl-code-ethics}{ACL Ethics Policy}. This work involves the use of publicly available datasets and does not include any personally identifiable information. For our human evaluation, we employed third-party evaluators, ensuring a rate of over \$20 per hour. An ethical concern worth mentioning is the use of language models for text generation, which may produce harmful content, either through intentional misuse by users or unintentionally due to the training data or algorithms. We declare that there are no conflicts of interest that could potentially influence the outcomes, interpretations, or conclusions of this research. All funding sources supporting this study are acknowledged in the acknowledgments section. We have diligently documented our methodology, experiments, and results, and we commit to sharing our code, data, and other relevant resources to enhance reproducibility and further advancements in the field.

\section*{Acknowledgments}
This work was partially supported by the MOE Liberal Arts and Social Sciences Foundation (No.23YJAZH210), Major Program of National Social Science Foundation (No.23\&ZD309), Henan Provincial Center for Outstanding Overseas Scientists (No.GZS2025004),  High Level Talent International Training Program of Henan Province (No.GCC2025010), Henan Provincial Science and Technology Project (No.252102210150), Key Scientific Research Project for Universities in Henan Province (No.25A520014), and the Chinese Scholarship Council (Grant No.202308410339). Moreover, Matthias Aßenmacher received funding from the Deutsche Forschungsgemeinschaft (DFG, German Research Foundation) as part of BERD@NFDI, under Grant No.460037581. Julian Rodemann acknowledges the funding support from the Federal Statistical Office of Germany within the co-operation project "Machine Learning in Official Statistics", the Bavarian Institute for Digital Transformation (bidt) and the Bavarian Academy of Sciences (BAdW) within a graduate scholarship. We also thank the support from the Munich Center for Machine Learning (MCML), and the Department of Statistics, LMU Munich. Last but not least, we thank all the anonymous reviewers and area chair(s) for their constructive suggestions and help throughout the review process. 


\bibliography{CRV_EMNLP}

\begin{thebibliography}{39}
\providecommand{\natexlab}[1]{#1}

\bibitem[{Ackley et~al.(1985)Ackley, Hinton, and Sejnowski}]{ackley1985learning}
David~H Ackley, Geoffrey~E Hinton, and Terrence~J Sejnowski. 1985.
\newblock A learning algorithm for boltzmann machines.
\newblock \emph{Cognitive science}, 9(1):147--169.

\bibitem[{Aichberger et~al.(2025)Aichberger, Schweighofer, Ielanskyi, and Hochreiter}]{aichberger2024semanticallydiverselanguagegeneration}
Lukas Aichberger, Kajetan Schweighofer, Mykyta Ielanskyi, and Sepp Hochreiter. 2025.
\newblock Improving uncertainty estimation through semantically diverse language generation.

\bibitem[{at~al.(2023)}]{llama2}
Hugo~Touvron at~al. 2023.
\newblock \href {https://arxiv.org/abs/2307.09288} {Llama 2: Open foundation and fine-tuned chat models}.
\newblock \emph{Preprint}, arXiv:2307.09288.

\bibitem[{Basu et~al.(2021)Basu, Ramachandran, Keskar, and Varshney.}]{basubadsampling}
Sourya Basu, Govardana~Sachitanandam Ramachandran, Nitish~Shirish Keskar, and Lav~R. Varshney. 2021.
\newblock Mirostat: a neural text decoding algorithm that directly controls perplexity.

\bibitem[{Bingham(1973)}]{bingham1973independent}
NH~Bingham. 1973.
\newblock Independent and stationary sequences of random variables.

\bibitem[{Carlsson et~al.(2025)Carlsson, Liu, Ward, Kurfali, and Nivre}]{carlsson2024hyperfittingphenomenonsharpeningstabilizing}
Fredrik Carlsson, Fangyu Liu, Daniel Ward, Murathan Kurfali, and Joakim Nivre. 2025.
\newblock The hyperfitting phenomenon: Sharpening and stabilizing llms for open-ended text generation.
\newblock In \emph{2025 International Conference on Learning Representations ({ICLR} 2025)}.

\bibitem[{Chen et~al.(2025)Chen, Hagrass, and Klusowski}]{chen2024decodinggameminimaxoptimality}
Sijin Chen, Omar Hagrass, and Jason~M. Klusowski. 2025.
\newblock Decoding game: On minimax optimality of heuristic text generation strategies.
\newblock In \emph{2025 International Conference on Learning Representations ({ICLR} 2025)}.

\bibitem[{{DeepSeek-AI}(2024)}]{deepseekai2024deepseekv3technicalreport}
{DeepSeek-AI}. 2024.
\newblock \href {https://arxiv.org/abs/2412.19437} {Deepseek-v3 technical report}.
\newblock \emph{Preprint}, arXiv:2412.19437.

\bibitem[{{Dubey et al.}(2024)}]{dubey2024llama3herdmodels}
{Dubey et al.} 2024.
\newblock \href {https://arxiv.org/abs/2407.21783} {The llama 3 herd of models}.
\newblock \emph{Preprint}, arXiv:2407.21783.

\bibitem[{Fan et~al.(2018)Fan, Lewis, and Dauphin}]{Fantopk}
Angela Fan, Mike Lewis, and Yann~N. Dauphin. 2018.
\newblock Hierarchical neural story generation.
\newblock \emph{Association for Computational Linguistics}, 11.

\bibitem[{Fang et~al.(2021)Fang, Zeng, Liu, Bo, Dong, and Chen}]{storygeneration}
Le~Fang, Tao Zeng, Chaochun Liu, Liefeng Bo, Wen Dong, and Changyou Chen. 2021.
\newblock Transformer-based conditional variational autoencoder for controllable story generation.
\newblock \emph{arXiv preprint arXiv:2101.00828, 2021}, 17.

\bibitem[{Freitag and Al-Onaizan(2017)}]{Freitag_2017}
Markus Freitag and Yaser Al-Onaizan. 2017.
\newblock \href {https://doi.org/10.18653/v1/w17-3207} {Beam search strategies for neural machine translation}.
\newblock In \emph{Proceedings of the First Workshop on Neural Machine Translation}. Association for Computational Linguistics.

\bibitem[{Garces~Arias et~al.(2025)Garces~Arias, Li, Heumann, and Assenmacher}]{garces-arias-etal-2025-decoding}
Esteban Garces~Arias, Meimingwei Li, Christian Heumann, and Matthias Assenmacher. 2025.
\newblock Decoding decoded: Understanding hyperparameter effects in open-ended text generation.
\newblock In \emph{Proceedings of the 31st International Conference on Computational Linguistics}, pages 9992--10020.

\bibitem[{Garces-Arias et~al.(2024)Garces-Arias, Rodemann, Li, Heumann, and Aßenmacher}]{estebanACS}
Esteban Garces-Arias, Julian Rodemann, Meimingwei Li, Christian Heumann, and Matthias Aßenmacher. 2024.
\newblock Adaptive contrastive search: Uncertainty-guided decoding for open-ended text generation.
\newblock In \emph{Findings of the Association for Computational Linguistics ({EMNLP} 2024)}, pages 15060--15080.

\bibitem[{Hewitt et~al.(2022)Hewitt, Manning, and Liang}]{hewitt-etal-2022-truncation}
John Hewitt, Christopher Manning, and Percy Liang. 2022.
\newblock \href {https://doi.org/10.18653/v1/2022.findings-emnlp.249} {Truncation sampling as language model desmoothing}.
\newblock In \emph{Findings of the Association for Computational Linguistics: EMNLP 2022}, pages 3414--3427, Abu Dhabi, United Arab Emirates. Association for Computational Linguistics.

\bibitem[{Holtzman et~al.(2020)Holtzman, Buys, Li~Du, and Choi}]{Holtzmandecode}
Ari Holtzman, Jan Buys, Maxwell~Forbes Li~Du, and Yejin Choi. 2020.
\newblock The curious case of neural text degeneration.
\newblock In \emph{2020 International Conference on Learning Representations ({ICLR} 2020)}.

\bibitem[{Jelinek et~al.(1977)Jelinek, Mercer, Bahl, and Baker}]{ppl}
Fred Jelinek, Robert~L Mercer, Lalit~R Bahl, and James~K Baker. 1977.
\newblock Perplexity—a measure of the difficulty of speech recognition tasks.
\newblock \emph{The Journal of the Acoustical Society of America}, 62(S1):S63--S63.

\bibitem[{{Jiang et al.}(2024)}]{jiang2023mistral7b}
{Jiang et al.} 2024.
\newblock \href {https://arxiv.org/abs/2310.06825} {Mistral 7b}.
\newblock \emph{Preprint}, arXiv:2310.06825.

\bibitem[{Leng et~al.(2024)Leng, Zhang, and Zhang}]{contextcompletion}
Siyi Leng, Zhenxin Zhang, and Liqiang Zhang. 2024.
\newblock A point contextual transformer network for point cloud completion.
\newblock \emph{Expert Systems with Applications}, 24.

\bibitem[{Li et~al.(2023)Li, Holtzman, Fried, Liang, Eisner, Hashimoto, Zettlemoyer, and Lewis}]{cd}
Xiang~Lisa Li, Ari Holtzman, Daniel Fried, Percy Liang, Jason Eisner, Tatsunori Hashimoto, Luke Zettlemoyer, and Mike Lewis. 2023.
\newblock Contrastive decoding: Open-ended text generation as optimization.
\newblock \emph{Association for Computational Linguistics}, 25.

\bibitem[{Meister et~al.(2023{\natexlab{a}})Meister, Pimentel, Wiher, and Cotterell.}]{typicalsampling}
Clara Meister, Tiago Pimentel, Gian Wiher, and Ryan Cotterell. 2023{\natexlab{a}}.
\newblock Locally typical sampling.
\newblock \emph{Transactions of the Association for Computational Linguistics}, 11:102--121.

\bibitem[{Meister et~al.(2023{\natexlab{b}})Meister, Pimentel, Wiher, and Cotterell}]{meister2023locallytypicalsampling}
Clara Meister, Tiago Pimentel, Gian Wiher, and Ryan Cotterell. 2023{\natexlab{b}}.
\newblock \href {https://arxiv.org/abs/2202.00666} {Locally typical sampling}.
\newblock \emph{Preprint}, arXiv:2202.00666.

\bibitem[{Merity et~al.(2017)Merity, Xiong, Bradbury, and Socher}]{wikitext1314}
Stephen Merity, Caiming Xiong, James Bradbury, and Richard Socher. 2017.
\newblock Pointer sentinel mixture models.
\newblock In \emph{5th International Conference on Learning Representations({ICLR} 2017)}.

\bibitem[{{Mesnard et al.}(2024)}]{gemma}
{Mesnard et al.} 2024.
\newblock \href {https://doi.org/10.48550/arXiv.2403.08295} {Gemma: Open models based on gemini research and technology}.
\newblock \emph{arXiv}, abs/2403.08295.

\bibitem[{Nguyen et~al.(2025{\natexlab{a}})Nguyen, Baker, Neo, Roush, Kirsch, and Shwartz-Ziv}]{nguyen2024turningheatminpsampling}
Minh Nguyen, Andrew Baker, Clement Neo, Allen Roush, Andreas Kirsch, and Ravid Shwartz-Ziv. 2025{\natexlab{a}}.
\newblock Turning up the heat: Min-p sampling for creative and coherent llm outputs.
\newblock In \emph{2025 International Conference on Learning Representations ({ICLR} 2025)}.

\bibitem[{Nguyen et~al.(2025{\natexlab{b}})Nguyen, Baker, Neo, Roush, Kirsch, and Shwartz-Ziv}]{nguyen2025turningheatminpsampling}
Minh Nguyen, Andrew Baker, Clement Neo, Allen Roush, Andreas Kirsch, and Ravid Shwartz-Ziv. 2025{\natexlab{b}}.
\newblock \href {https://arxiv.org/abs/2407.01082} {Turning up the heat: Min-p sampling for creative and coherent llm outputs}.
\newblock \emph{Preprint}, arXiv:2407.01082.

\bibitem[{{Pillutla at al.}(2021)}]{mauve}
{Pillutla at al.} 2021.
\newblock Mauve: Measuring the gap between neural text and human text using divergence frontiers.
\newblock In \emph{Annual Conference on Neural Information Processing Systems ({NeurIPS} 2021)}, pages 4816--4828.

\bibitem[{Qwen(2024)}]{qwen2.5}
Qwen. 2024.
\newblock \href {https://qwenlm.github.io/blog/qwen2.5/} {Qwen2.5: A party of foundation models}.

\bibitem[{Radford et~al.(2019)Radford, Wu, Child, Luan, Amodei, and Sutskever}]{radford2019gpt}
Alec Radford, Jeff Wu, Rewon Child, David Luan, Dario Amodei, and Ilya Sutskever. 2019.
\newblock Language models are unsupervised multitask learners.

\bibitem[{Su et~al.(2022)Su, Lan, Wang, Yogatama, Kong, and Collier}]{su2022contrastive}
Yixuan Su, Tian Lan, Yan Wang, Dani Yogatama, Lingpeng Kong, and Nigel Collier. 2022.
\newblock A contrastive framework for neural text generation.
\newblock In \emph{Annual Conference on Neural Information Processing Systems ({NeurIPS} 2022)}.

\bibitem[{Su et~al.(2021)Su, Wang, Cai, Baker, Korhonen, , and Collier}]{dustysteam}
Yixuan Su, Yan Wang, Deng Cai, Simon Baker, Anna Korhonen, , and Nigel Collier. 2021.
\newblock Prototype-to-style: dialogue generation with style-aware editing on retrieval memory.
\newblock \emph{IEEE/ACM Transactions on Audio, Speech, and Language Processing}, 29.

\bibitem[{{Vaswani et al.}(2017)}]{vaswani2017attention}
{Vaswani et al.} 2017.
\newblock Attention is all you need.
\newblock \emph{Advances in neural information processing systems}, 30.

\bibitem[{Vijayakumar et~al.(2018)Vijayakumar, Cogswell, Selvaraju, Qing~Sun, Crandall, and Batra}]{Ashwindecode}
Ashwin~K Vijayakumar, Michael Cogswell, Ramprasath~R. Selvaraju, Stefan~Lee Qing~Sun, David Crandall, and Dhruv Batra. 2018.
\newblock Diverse beam search: Decoding diverse solutions from neural sequence models.
\newblock \emph{Association for the Advancement of Artificial Intelligence}, 16.

\bibitem[{Welleck et~al.(2020)Welleck, Kulikov, Kim, Pang, and Cho}]{semanticinconsistency}
Sean Welleck, Ilia Kulikov, Jaedeok Kim, Richard~Yuanzhe Pang, and Kyunghyun Cho. 2020.
\newblock Consistency of a recurrent language model with respect to incomplete decoding.
\newblock In \emph{Proceedings of the 2020 Conference on Empirical Methods in Natural Language Processing ({EMNLP} 2020)}, pages 5553--5568.

\bibitem[{Yu et~al.(2024)Yu, Kairouz, Oh, and Xu}]{yu2024privacypreservinginstructionsaligninglarge}
Da~Yu, Peter Kairouz, Sewoong Oh, and Zheng Xu. 2024.
\newblock \href {https://arxiv.org/abs/2402.13659} {Privacy-preserving instructions for aligning large language models}.
\newblock \emph{Preprint}, arXiv:2402.13659.

\bibitem[{Zhang et~al.(2020)Zhang, Kishore, Wu, Weinberger, and Artzi}]{bert-score}
Tianyi Zhang, Varsha Kishore, Felix Wu, Kilian~Q. Weinberger, and Yoav Artzi. 2020.
\newblock Bertscore: Evaluating text generation with bert.
\newblock In \emph{2020 International Conference on Learning Representations({ICLR} 2020)}.

\bibitem[{{Zhang et al.}(2022)}]{zhang2022opt}
{Zhang et al.} 2022.
\newblock \href {https://arxiv.org/abs/2205.01068} {Opt: Open pre-trained transformer language models}.
\newblock \emph{Preprint}, arXiv:2205.01068.

\bibitem[{Zhu et~al.(2024)Zhu, Hao, He, Ai, and Wang}]{zhu2024improving}
Wenhong Zhu, Hongkun Hao, Zhiwei He, Yiming Ai, and Rui Wang. 2024.
\newblock Improving open-ended text generation via adaptive decoding.
\newblock \emph{arXiv preprint arXiv:2402.18223}.

\bibitem[{{Zhu at al.}(2015)}]{book1974}
{Zhu at al.} 2015.
\newblock Aligning books and movies: Towards story-like visual explanations by watching movies and reading books.
\newblock In \emph{2015 {IEEE} International Conference on Computer Vision ({ICCV} 2015)}, pages 19--27.

\end{thebibliography}

\clearpage

\onecolumn

\appendix

\section*{Appendix}


\section{Proofs}\label{app:A}

For ease of exposition, we restate all claims before proving them.

\subsection{Proof of Proposition~1}

\textbf{Proposition} (Unbiasedness of \(H_{\text{glob}}\))
    Assume that the process \(\{H(X)_{t}\}_{t\ge 1}\) is stationary. Then, 
    \[
\mathbb E\bigl[H_{\text{glob}}\bigr] = \mathbb E[ H(X)_{t}]
\]
for any $\lambda \in [0,1)$.

\begin{proof}
    Define the normalized weights
\[
w_i = \frac{\lambda^{(t-i)}}{\sum_{s=1}^{t} \lambda^{(t-s)}}, \quad \text{such that} \quad \sum_{i=1}^{t} w_i = 1,
\]
for any $\lambda \in [0,1)$.
Then we can write the estimator \eqref{eq:hglob} as
\[
H_{\text{glob}} = \sum_{i=1}^{t} w_i\, H(X)_i.
\]
Taking expectation and using linearity together with the stationarity assumption, we obtain
\[
\mathbb E\bigl[H_{\text{glob}}\bigr] = \sum_{i=1}^{t} w_i\, \mathbb E\bigl[H(X)_i\bigr]
= \sum_{i=1}^{t} w_i\, H_0
= H_0.
\]
Since under stationarity, the instantaneous entropy's expectation at $t$ is
$
\mathbb E[H(X)_t] = H_0,
$
the claim immediately follows:
\[
E\bigl[H_{\text{glob}}\bigr] = \mathbb E[H(X)_t].
\]
\hfill $\square$
\end{proof}

\subsection{Proof of Proposition~2}

\begin{proof}
We show that the variance of \(H_{\text{glob}}\) vanishes as \(t \to \infty\), so that by Chebyshev's inequality the estimator converges in probability to \( \mathbb E[H(X)_t]\) under the following assumptions
\begin{enumerate}
    \item[\textbf{(1)}] \textbf{Stationarity and Ergodicity:} The sequence \(\{H(X)_t\}_{t\ge 1}\) is stationary and ergodic with
    \[
        \forall t \in \{1, \dots, T\}: \;  E\Bigl[H(X)_t\Bigr] =: H_0.
    \]

    \item[\textbf{(2)}] \textbf{Boundedness:} There exists a constant \(M < \infty\) such that
    \[
        |H(X)_t| \le M \quad \text{for all } t.
    \]

    \item[\textbf{(3)}] \textbf{Decaying Covariances:} The covariances \(\operatorname{Cov}\bigl(H(X)_t,H(X)_{s}\bigr)\) decay sufficiently fast as \(|t-s|\) increases.
\end{enumerate}

Decaying covariances can be ensured by, e.g., strong mixing of the sequence \(\{H(X)_t\}\) with mixing coefficients \(\alpha(n)\) satisfying
        \[
        \sum_{n=1}^{\infty} \alpha(n)^{\delta/(2+\delta)} < \infty
        \]

    for some \(\delta > 0\). 

The argument is as follows. Define the normalized weights
    \[
    w_i = \frac{\lambda^{t-i}}{\sum_{s=1}^{t} \lambda^{t-s}}, \quad \text{such that} \quad \sum_{i=1}^{t} w_i = 1,
    \]

for any $\lambda \in [0,1)$.
    
Since
    \[
    H_{\text{glob}, t} = \sum_{i=1}^{t} w_i\, H(X)_t,
    \]

its variance is given by
    \[
\operatorname{Var}\bigl(H_{\text{glob}, t}\bigr) = \sum_{i=1}^{t}\sum_{s=1}^{t} w_i\, w_s\, \operatorname{Cov}\Bigl(H(X)_i, H(X)_{s}\Bigr).
    \]

 Per assumptions, we have that all \(H(X)_t\) are bounded by \(M\) and their covariances decay sufficiently fast as \(|t-s|\) increases. Moreover, the weights form a convex combination. Under standard mixing conditions (see, e.g., \citet{bingham1973independent}), one can show that
    \[
    \operatorname{Var}\bigl(H_{\text{glob}, t}\bigr) = \mathcal O\Bigl(\frac{1}{t}\Bigr),
    \]

where $\mathcal O$ is the Landau (\say{big O}) notation.
Then, by Chebyshev's inequality, for any \(\epsilon>0\) we have
    \[
    \mathbb P\Bigl(\bigl|H_{\text{glob}} - \mathbb E[H_{\text{glob}}]\bigr| > \epsilon\Bigr) \le \frac{\operatorname{Var}\bigl(H_{\text{glob}}\bigr)}{\epsilon^2} \to 0 
    \]
where  $\text{as } t \to \infty$.

Together with Proposition 1, this implies
    \[
    H_{\text{glob},t} \stackrel{\mathbb P}{\longrightarrow} \mathbb E[H(X)_t]
    \]

as $t \rightarrow \infty$. Thus, \(H_{\text{glob},t}\) is a consistent estimator.
\end{proof}


\section{Further (Design Choice) Experiments}
\label{app:B}

\noindent\textbf{Optimal $\lambda$ Selection.} We further investigate how different values of $\lambda$ influence the quality of text generations. We find that model performance remains stable for $\lambda \in \{0.91, 0.93, 0.95, 0.97, 0.99\}$, thus we select $\lambda = 0.95$ as our default parameter. This experiment is based on the GPT-2 XL model \citep{radford2019gpt} to balance computational expenses. 

\begin{table}[ht]
\centering
\resizebox{1\textwidth}{!}{
\begin{tabular}{cccccccccc}
\toprule
\multirow{2}{*}{ Method } & \multicolumn{3}{c}{ Wikitext } & \multicolumn{3}{c}{ Wikinews } & \multicolumn{3}{c}{ BookCorpus } \\
\cmidrule(lr){2-4} \cmidrule(lr){5-7} \cmidrule(lr){8-10}
 & div. (\%) & MAUVE (\%) & coh. & div. (\%) & MAUVE (\%) & coh.  & div. (\%) & MAUVE (\%) & coh. \\
\midrule
$\lambda=0.91$ & 95.25 & 83.64 & -2.60 & 95.86 & 89.39 & -2.34 & 96.41 & 83.43 & -2.61 \\
$\lambda=0.93$ & 95.22 & 81.39 & -2.60 & 96.31 & 92.79 & -2.34 & 96.39 & 81.55 & -2.61 \\
$\lambda=0.95$ & 95.64 & 85.55 & -2.62 & 96.19 & 90.61 & -2.35 & 96.35 & 85.83 & -2.62 \\
$\lambda=0.97$ & 96.00 & 79.27 & -2.63 & 96.28 & 92.32 & -2.36 & 96.60 & 82.79 & -2.63 \\
$\lambda=0.99$ & 95.95 & 80.48 & -2.67 & 96.46 & 90.35 & -2.39 & 96.75 & 79.73 & -2.65 \\

\bottomrule
\end{tabular}
}
\caption{Evaluation of performance across three datasets for $\lambda$ values in (0.91, 0.99).}
\label{tab:lambda_selection}
\end{table}

\noindent\textbf{Median Aggregation and Arctanh.} To validate our design choices of Median aggregation and Arctanh mapping, we conducted statistical significance testing across multiple aggregation and mapping combinations. Bootstrap analysis with 95\% confidence intervals demonstrated that our chosen Median + Arctanh method achieves significantly superior MAUVE scores (93.60 [92.71, 94.48]) compared to all alternative approaches ($p < 0.001, t = 4.98$). The effect sizes for these MAUVE improvements are consistently large (all Cohen's $d > 1.75$), with percentage improvements ranging from 3.61\% to 6.39\% over alternatives. While other combinations achieve better Diversity scores and certain alternatives show marginal improvements in Coherence, we follow common practices to guide these choices \citep{hewitt-etal-2022-truncation, zhu2024improving, nguyen2025turningheatminpsampling, yu2024privacypreservinginstructionsaligninglarge, meister2023locallytypicalsampling}. The Friedman test across all metrics ($\chi^2 = 9.10$) indicates no statistically significant differences in overall performance across methods, suggesting that trade-offs exist between optimization targets.

\begin{table}[H]
\centering
\resizebox{1\textwidth}{!}{
\begin{tabular}{cccccc}
\toprule
\textbf{Aggregation} & \textbf{Mapping} & \textbf{Diversity} & \textbf{MAUVE} & \textbf{Coherence} & \textbf{Notes}\\
\midrule
Winsorized mean & Logarithmic map & 95.22 & 88.44 & -2.42 & -\\
Median & Logarithmic map & 95.71 & 90.16 & -2.48 & - \\
Winsorized mean & Arctanh & 94.49 & 87.98 & -2.35 & - \\
Mean & Logarithmic map & 95.75 & 88.91 & -2.35 & - \\
Mean & Arctanh & 97.15 & 90.34 & -2.61 & - \\
Median & Arctanh & 95.20 & 93.60 & -2.38 & GUARD (Ours)\\

\bottomrule
\end{tabular}
}
\caption{Evaluation of different value assignment strategies and function mappings using the Wikinews dataset.}
\label{tab:median_selection}
\end{table}


\noindent\textbf{Optimal Locality Window Selection.} We evaluated the performance of various locality window sizes, $w \in \{2, 3, 4, 5, 6, 7, 8, 9\}$, with the results summarized in Table \ref{tab:ablation_selectw}. The analysis indicates that a window size of $w = 7$ yields good performance, while other choices in that range also display similar scores. To further validate this finding, we conducted additional experiments using multiple models, including Qwen2.5-7B, Llama2, Deepseek-llm-7B-base, Llama-3, Mistral-v0.3, Gemma-7B, and GPT2-xl. Detailed results for these experiments are provided in Tables \ref{tab:ablation_selectw},  \ref{tab:ablation_selectw_llama2}, \ref{tab:ablation_selectw_deepseek}, \ref{tab:ablation_selectw_llama3}, \ref{tab:ablation_selectw_mirstral}, \ref{tab:ablation_selectw_gemma}, and \ref{tab:ablation_selectw_gpt2}.

\begin{table*}[!ht]
\centering

\resizebox{1\textwidth}{!}{
\begin{tabular}{cccccccccc}
\toprule
\multirow{2}{*}{Method} & \multicolumn{3}{c}{Wikitext} & \multicolumn{3}{c}{Wikinews} & \multicolumn{3}{c}{BookCorpus} \\
\cmidrule(lr){2-4} \cmidrule(lr){5-7} \cmidrule(lr){8-10}
 & div. (\%) & MAUVE (\%) & coh. & div. (\%) & MAUVE (\%) & coh. & div. (\%) & MAUVE (\%) & coh. (\%) \\
\midrule
GUARD (w=2) & 93.56 & 87.89 & -2.57 & 94.64 & 94.20 & -2.35 & 95.94 & 91.25 & -2.52 \\
GUARD (w=3) & 94.06 & 88.87 & -2.67 & 96.17 & 94.29 & -2.41 & 96.89 & 90.43 & -2.61 \\
GUARD (w=4) & 93.11 & 89.32 & -2.57 & 94.73 & 94.83 & -2.32 & 95.55 & 93.58 & -2.46 \\
GUARD (w=5) & 92.97 & 87.26 & -2.61 & 96.05 & 93.56 & -2.39 & 96.51 & 92.30 & -2.54 \\
GUARD (w=6) & 92.92 & 86.85 & -2.57 & 96.05 & 93.56 & -2.34 & 95.69 & 93.57 & -2.45 \\
GUARD (w=7) & 92.86 & 90.82 & -2.61 & 95.20 & 93.60 & -2.38 & 96.18 & 92.59 & -2.52 \\
GUARD (w=8) & 92.51 & 88.29 & -2.57 & 94.45 & 93.66 & -2.32 & 95.48 & 91.56 & -2.46 \\
GUARD (w=9) & 92.82 & 89.78 & -2.61 & 94.94 & 94.53 & -2.38 & 96.14 & 89.96 & -2.52 \\
\bottomrule
\end{tabular}
}

\caption{Qwen2.5-7B: The performance under different datasets and $w$ values shows that $w$ = 7 is an appropriate locality window. As the value of $w$ increases, the data value decreases; hence, no further experiments are conducted.
}
\label{tab:ablation_selectw}
\end{table*}

\begin{table*}[!ht]
\centering

\resizebox{1\textwidth}{!}{
\begin{tabular}{cccccccccc}
\toprule
\multirow{2}{*}{Method} & \multicolumn{3}{c}{Wikitext} & \multicolumn{3}{c}{Wikinews} & \multicolumn{3}{c}{ BookCorpus } \\
\cmidrule(lr){2-4} \cmidrule(lr){5-7} \cmidrule(lr){8-10}
 & div. (\%) & MAUVE (\%) & coh. & div. (\%) & MAUVE (\%) & coh. & div. (\%) & MAUVE (\%) & coh. \\
\midrule
GUARD (w=2) & 95.49 & 90.59 & -2.67 & 96.19 & 93.21 & -2.25 & 96.21 & 93.08 & -2.62 \\
GUARD (w=3) & 96.73 & 91.46 & -2.87 & 97.18 & 95.33 & -2.36 & 97.10 & 92.76 & -2.61 \\
GUARD (w=4) & 94.73 & 91.27 & -2.64 & 96.20 & 92.90 & -2.26 & 96.89 & 91.97 & -2.64 \\
GUARD (w=5) & 95.82 & 87.91 & -2.77 & 96.91 & 92.85 & -2.33 & 95.59 & 92.46 & -2.59 \\
GUARD (w=6) & 94.50 & 90.00 & -2.66 & 96.45 & 93.41 & -2.24 & 96.15 & 92.78 & -2.71 \\
GUARD (w=7) & 95.60 & 89.52 & -2.75 & 96.76 & 93.34 & -2.31 & 96.21 & 93.08 & -2.62 \\
GUARD (w=8) & 94.74 & 87.78 & -2.66 & 96.48 & 90.41 & -2.25 & 96.10 &  92.78 & -2.49 \\
GUARD (w=9) & 95.38 & 90.78 & -2.74 & 96.54 & 94.12 & -2.31 & 95.98 & 92.12 & -2.59 \\
\bottomrule
\end{tabular}}

\caption{ Llama-2: The performance under different datasets and $w$ values shows that $w$ = 7 is the appropriate locality window. 
}
\label{tab:ablation_selectw_llama2}
\end{table*}

\begin{table*}[!ht]
\centering
\resizebox{1\textwidth}{!}{
\begin{tabular}{cccccccccc}
\toprule
\multirow{2}{*}{Method} & \multicolumn{3}{c}{Wikitext} & \multicolumn{3}{c}{Wikinews} & \multicolumn{3}{c}{ BookCorpus } \\
\cmidrule(lr){2-4} \cmidrule(lr){5-7} \cmidrule(lr){8-10}
 & div. (\%) & MAUVE (\%) & coh. & div. (\%) & MAUVE (\%) & coh. & div. (\%) & MAUVE (\%) & coh. \\
\midrule
GUARD (w=2) & 95.62 & 89.45 & -2.65 & 96.56 & 96.50 & -2.21 & 96.32 & 90.36 & -2.58 \\
GUARD (w=3) & 97.21 & 88.12 & -2.74 & 97.55 & 93.36 & -2.33 & 97.34 & 91.93 & -2.68 \\
GUARD (w=4) & 95.41 & 84.49 & -2.54 & 96.64 & 94.31 & -2.20 & 96.09 & 90.43 & -2.50 \\
GUARD (w=5) & 96.82 & 89.74 & -2.68 & 97.24 & 94.32 & -2.29 & 97.02 & 89.17 & -2.61 \\
GUARD (w=6) & 95.64 & 83.35 & -2.48 & 96.59 & 93.77 & -2.20 & 96.34 & 89.46 & -2.52 \\
GUARD (w=7) & 96.55 & 88.96 & -2.65 & 97.14 & 94.84 & -2.23 & 96.86 & 90.07 & -2.61 \\
GUARD (w=8) & 95.70 & 91.09 & -2.56 & 96.71 & 94.27 & -2.21 & 96.23 & 88.95 & -2.52 \\
GUARD (w=9) & 96.52 & 85.69 & -2.65 & 97.08 & 94.24 & -2.27 & 96.90 & 91.80 & -2.61 \\
\bottomrule
\end{tabular}
}
\caption{ Deepseek-llm-7B-base: The performance under different datasets and $w$ values shows that $w$ = 7 is the appropriate locality window. 
}
\label{tab:ablation_selectw_deepseek}
\end{table*}

\begin{table*}[!ht]
\centering
\resizebox{1\textwidth}{!}{
\begin{tabular}{cccccccccc}
\toprule
\multirow{2}{*}{Method} & \multicolumn{3}{c}{Wikitext} & \multicolumn{3}{c}{Wikinews} & \multicolumn{3}{c}{ BookCorpus } \\
\cmidrule(lr){2-4} \cmidrule(lr){5-7} \cmidrule(lr){8-10}
 & div. (\%) & MAUVE (\%) & coh. & div. (\%) & MAUVE (\%) & coh. & div. (\%) & MAUVE (\%) & coh. \\
\midrule
GUARD (w=2) & 95.47 & 89.85 & -2.57 & 96.16 & 94.88 & -2.15 & 96.02 & 91.20 & -2.49 \\
GUARD (w=3) & 96.65 & 89.74 & -2.66 & 97.25 & 95.49 & -2.24 & 96.84 & 91.52 & -2.57 \\
GUARD (w=4) & 94.94 & 89.98 & -2.45 & 95.82 & 96.12 & -2.12 & 95.75 & 90.13 & -2.41 \\
GUARD (w=5) & 95.95 & 88.72 & -2.56 & 96.67 & 93.44 & -2.21 & 96.39 & 92.50 & -2.52 \\
GUARD (w=6) & 94.88 & 90.74 & -2.47 & 95.58 & 94.51 & -2.14 & 95.73 & 91.14 & -2.44 \\
GUARD (w=7) & 95.84 & 91.25 & -2.55 & 96.45 & 94.94 & -2.20 & 96.21 & 92.09 & -2.50 \\
GUARD (w=8) & 94.77 & 90.92 & -2.47 & 95.95 & 93.90 & -2.15 & 95.82 & 91.11 & -2.43 \\
GUARD (w=9) & 95.62 & 89.44 & -2.55 & 96.59 & 92.31 & -2.20 & 96.04 & 91.50 & -2.48 \\
\bottomrule
\end{tabular}
}
\caption{ Llama-3: The performance under different datasets and $w$ values shows that $w$ = 7 is the appropriate locality window. 
}
\label{tab:ablation_selectw_llama3}
\end{table*}

\begin{table*}[!ht]
\centering
\resizebox{1\textwidth}{!}{
\begin{tabular}{cccccccccc}
\toprule
\multirow{2}{*}{Method} & \multicolumn{3}{c}{Wikitext} & \multicolumn{3}{c}{Wikinews} & \multicolumn{3}{c}{ BookCorpus } \\
\cmidrule(lr){2-4} \cmidrule(lr){5-7} \cmidrule(lr){8-10}
 & div. (\%) & MAUVE (\%) & coh. & div. (\%) & MAUVE (\%) & coh. & div. (\%) & MAUVE (\%) & coh. \\
\midrule
GUARD (w=2) & 94.87 & 92.52 & -2.58 & 96.68 & 91.68 & -2.20 & 96.25 & 90.89 & -2.60 \\
GUARD (w=3) & 96.57 & 93.39 & -2.71 & 97.49 & 91.89 & -2.31 & 97.29 & 91.02 & -2.70 \\
GUARD (w=4) & 95.10 & 91.64 & -2.53 & 96.73 & 92.45 & -2.19 & 96.10 & 91.26 & -2.54 \\
GUARD (w=5) & 95.70 & 91.43 & -2.66 & 97.32 & 89.66 & -2.29 & 96.82 & 91.52 & -2.65 \\
GUARD (w=6) & 94.97 & 91.14 & -2.54 & 96.61 & 91.50 & -2.21 & 96.20 & 91.57 & -2.55 \\
GUARD (w=7) & 95.69 & 90.58 & -2.63 & 96.85 & 90.49 & -2.27 & 96.78 & 94.56 & -2.64 \\
GUARD (w=8) & 95.05 & 89.26 & -2.21 & 96.56 & 92.41 & -2.21 & 96.20 & 92.64 & -2.57 \\
GUARD (w=9) & 95.57 & 92.42 & -2.63 & 97.07 & 93.88 & -2.26 & 96.65 & 89.73 & -2.62 \\
\bottomrule
\end{tabular}
}
\caption{ Mistral-v0.3: The performance under different datasets and $w$ values shows that $w$ = 7 is the appropriate locality window. 
}
\label{tab:ablation_selectw_mirstral}
\end{table*}

\begin{table*}[!ht]
\centering
\resizebox{1\textwidth}{!}{
\begin{tabular}{cccccccccc}
\toprule
\multirow{2}{*}{Method} & \multicolumn{3}{c}{Wikitext} & \multicolumn{3}{c}{Wikinews} & \multicolumn{3}{c}{ BookCorpus } \\
\cmidrule(lr){2-4} \cmidrule(lr){5-7} \cmidrule(lr){8-10}
& div. (\%) & MAUVE (\%) & coh. & div. (\%) & MAUVE (\%) & coh. & div. (\%) & MAUVE (\%) & coh. \\
\midrule
GUARD (w=2) & 95.32 & 89.45 & -2.65 & 96.71 & 87.45 & -2.16 & 96.28 & 89.39 & -2.35 \\
GUARD (w=3) & 96.81 & 90.16 & -2.62 & 97.53 & 91.44 & -2.27 & 95.46 & 90.41 & -2.43 \\
GUARD (w=4) & 95.86 & 88.22 & -2.47 & 96.60 & 89.90 & -2.14 & 96.88 & 89.43 & -2.56 \\
GUARD (w=5) & 96.65 & 88.47 & -2.56 & 97.33 & 89.22 & -2.22 & 95.26 & 91.65 & -2.67 \\
GUARD (w=6) & 95.46 & 90.56 & -2.46 & 96.70 & 92.74 & -2.15 & 96.44 & 90.16 & -2.52 \\
GUARD (w=7) & 96.54 & 88.16 & -2.55 & 97.23 & 90.76 & -2.22 & 96.97 & 91.29 & -2.65 \\
GUARD (w=8) & 95.59 & 87.81 & -2.49 & 96.71 & 91.62 & -2.15 & 96.30 & 91.09 &  -2.35\\
GUARD (w=9) & 96.20 & 89.79 & -2.54 & 97.09 & 88.50 & -2.22 & 95.69 & 90.98 & -2.64 \\
\bottomrule
\end{tabular}
}
\caption{ Gemma-7B: The performance under different datasets and $w$ values shows that $w$ = 7 is the appropriate locality window. 
}
\label{tab:ablation_selectw_gemma}
\end{table*}

\clearpage

\begin{table*}[ht]
\centering
\resizebox{1\textwidth}{!}{
\begin{tabular}{cccccccccc}
\toprule
\multirow{2}{*}{Method} & \multicolumn{3}{c}{Wikitext} & \multicolumn{3}{c}{Wikinews} & \multicolumn{3}{c}{ BookCorpus } \\
\cmidrule(lr){2-4} \cmidrule(lr){5-7} \cmidrule(lr){8-10}
 & div. (\%) & MAUVE (\%) & coh. & div. (\%) & MAUVE (\%) & coh. & div. (\%) & MAUVE (\%) & coh. \\
\midrule
GUARD (w=2) & 96.64 & 84.21 & -2.79 & 96.79 & 91.26 & -2.48 & 97.29 & 84.71 & -2.76 \\
GUARD (w=3) & 96.64 & 84.21 & -2.79 & 96.79 & 91.26 & -2.48 & 97.29 & 84.71 & -2.76 \\
GUARD (w=4) & 94.39 & 83.56 & -2.49 & 95.86 & 87.44 & -2.25 & 95.56 & 83.13 & -2.51 \\
GUARD (w=5) & 96.28 & 83.06 & -2.66 & 96.49 & 89.83 & -2.36 & 96.66 & 83.37 & -2.64 \\
GUARD (w=6) & 94.60 & 79.72 & -2.50 & 95.92 & 91.78 & -2.26 & 95.69 & 81.43 & -2.52 \\
GUARD (w=7) & 95.64 & 85.55 & -2.62 & 96.19 & 90.61 & -2.35 & 96.35 & 85.83 & -2.62 \\
GUARD (w=8) & 94.69 & 84.25 & -2.53 & 95.50 & 90.60 & -2.55 & 95.68 & 82.50 & -2.53 \\
GUARD (w=9) & 95.56 & 78.64 & -2.62 & 96.17 & 89.35 & -2.35 & 96.24 & 80.30 & -2.61 \\
\bottomrule
\end{tabular}}
\caption{ GPT2-xl: The performance under different datasets and $w$ values shows that $w$ = 7 is the appropriate locality window. 
}
\label{tab:ablation_selectw_gpt2}
\end{table*}


\section{Pseudo Code Design}
\label{app:C}

{\small
\renewcommand{\algorithmicrequire}{\textbf{Input:}}  
\renewcommand{\algorithmicensure}{\textbf{Output:}}

\makeatletter
\newenvironment{breakablealgorithm}
  {
   \begin{center}
     \refstepcounter{algorithm}
     \hrule height.8pt depth0pt \kern2pt
     \renewcommand{\caption}[2][\relax]{
       {\raggedright\textbf{\ALG@name~\thealgorithm} ##2\par}%
       \ifx\relax##1\relax 
         \addcontentsline{loa}{algorithm}{\protect\numberline{\thealgorithm}##2}%
       \else 
         \addcontentsline{loa}{algorithm}{\protect\numberline{\thealgorithm}##1}%
       \fi
       \kern2pt\hrule\kern2pt
     }
  }{
     \kern2pt\hrule\relax
   \end{center}
  }
\makeatother

\begin{breakablealgorithm}
\caption{GUARD Algorithm}
\begin{algorithmic}[1]
    \Require \Statex
        Prompt, Model, MaxTokens, $w$ (window size), $\lambda$ (decay factor) 
    \Ensure \Statex
        Generated text continuation 
    \Function{GENERATETEXT}{Prompt, Model, MaxTokens, w, $\lambda$}
        \State $Output \gets \text{Prompt}$
        \State $entropy\_history \gets [\,]$; \quad $tokencounts \gets \{\}$; \quad $\varepsilon \gets 1e-6$
        \For{$t=1 \text{ to } MaxTokens$}
            \State $distribution \gets \text{Model.getDistribution(Output)}$
            \State $V \gets \text{distribution.size}$
            \State $H_{loc} \gets -\sum(\text{distribution} * \log(\text{distribution}))$  // Eq. 2
            \State $entropy\_history[t] \gets H_{loc}$
            \If{$t < w$}
                \State $\delta_{global} \gets \text{arctanh}((H_{loc} - \text{med}(entropy\_history)) / \log\lvert V \rvert)$ //Eq. 9
                \State $\delta_{loc} \gets 0$
            \Else
                \State $T \gets \text{len}(entropy\_history)$
                \State $numerator \gets \sum_{i=1}^{T} (\lambda^{T-i} * entropy\_history[i])$
                \State $denominator \gets \sum_{i=1}^{T} (\lambda^{T-i})$
                \State $H_{glob,T} \gets numerator / denominator$ //Eq. 1
                \State $med\_H\_recent \gets \text{median}(entropy\_history[T-w : T])$
                \State $med\_diff \gets med\_H\_recent - \text{med}(H_{glob,T})$
                \State $r_{change} \gets \lvert H_{loc} - entropy\_history[T-1] \rvert / (entropy\_history[T-1] + \varepsilon)$
                \State $r_{diff} \gets \lvert med\_diff \rvert / (\text{med}(H_{glob,T}) + \varepsilon)$
                \State $q \gets 1.0 + r_{change} + r_{diff}$ //Appendix \ref{app:B}
                \State $\delta_{loc} \gets q * \text{arctanh}((H_{loc} - med\_H\_recent) / \log\lvert V \rvert)$ //Eq. 8
                \State $\delta_{global} \gets q * \text{arctanh}(med\_diff / \log\lvert V \rvert)$ //Eq. 9
            \EndIf
            \State $\lambda_k \gets \lvert \delta_{loc} \rvert / (\lvert \delta_{loc} \rvert + \lvert \delta_{global} \rvert + \varepsilon)$ //Eq. 10
            \State $k_{signal} \gets \exp(\lambda_k * \delta_{loc} + (1 - \lambda_k) * \delta_{global})$
            \State $k_t \gets 10 * (1 - 1 / (k_{signal} + 1)) + 5$ //Eq. 7
            \State $k_{signal\_\alpha} \gets \exp(\lambda_k \delta_{loc} \log\lvert k_t \rvert + (1 - \lambda_k) \delta_{global} \log\lvert k_t \rvert)$
            \State $\alpha_t \gets 1 - 1 / (k_{signal\_\alpha} + 1)$ //Eq. 12
            \State $candidates \gets \text{Select top-}k_t \text{ candidates}$
            \State $scores \gets \{\}$
            \For{each token $v$ in candidates}
                \State $count \gets \text{tokencounts.get}(v, 0)$
                \State $scores[v] \gets \text{distribution}[v] * (\alpha_t^{\text{count}})$ //Eq. 11
            \EndFor
            \State $next\_token \gets \text{argmax}(scores)$
            \State $tokencounts[next\_token] \gets \text{tokencounts.get}(next\_token, 0) + 1$
            \State $Output \gets Output + next\_token$
        \EndFor
        \State \Return{$Output[\text{len(Prompt):}]$}
    \EndFunction
\end{algorithmic}
\end{breakablealgorithm}
}
\clearpage


\section{Metrics}
\label{app:D}

\subsection*{Diversity} This metric aggregates $\mathrm{n}$-gram repetition rates: $$\text{DIV}=\prod_{n=2}^4 \frac{\mid \text { unique } \mathrm{n} \text {-grams }\left(\mathrm{x}_{\text {cont }}\right) \mid}{\text { total } \mathrm{n} \text {-grams }\left(\mathrm{x}_{\text {cont }}\right) \mid}.$$ A low diversity score suggests the model suffers from repetition, and a high diversity score means the model-generated text is lexically diverse.

\subsection*{MAUVE} MAUVE \cite{mauve} is a metric designed to quantify how closely a model distribution \(Q\) matches a target distribution \(P\) of human texts. Two main types of error contribute to any discrepancy between \(Q\) and \(P\): 
\begin{itemize}
    \item \textbf{Type I Error:} \(Q\) assigns high probability to text that is unlikely under \(P\).
    \item \textbf{Type II Error:} \(Q\) fails to generate text that is plausible under \(P\).
\end{itemize}

These errors can be formalized using the Kullback--Leibler (KL) divergences \(\mathrm{KL}(Q \!\parallel\! P)\) and \(\mathrm{KL}(P \!\parallel\! Q)\). If \(P\) and \(Q\) do not share the same support, at least one of these KL divergences will be infinite. To address this issue, \citet{mauve} propose measuring errors through a mixture distribution
\[
R_{\lambda} \;=\; \lambda P \;+\; (1-\lambda)\,Q
\quad\text{with}\;\lambda \in (0,1).
\]
This leads to redefined Type I and Type II errors given by
\[
\mathrm{KL}\!\bigl(Q \!\parallel\! R_{\lambda}\bigr)
\quad \text{and} \quad
\mathrm{KL}\!\bigl(P \!\parallel\! R_{\lambda}\bigr),
\]
respectively.

\bigskip
\noindent
By varying \(\lambda\) and computing these two errors, one obtains a \emph{divergence curve}
\[
\mathcal{C}(P, Q)
\;=\;
\left\{
\left(
\exp\!\bigl(-c\,\mathrm{KL}(Q \!\parallel\! R_{\lambda})\bigr),\;
\exp\!\bigl(-c\,\mathrm{KL}(P \!\parallel\! R_{\lambda})\bigr)
\right)
\,:\,
R_\lambda = \lambda P + (1-\lambda) Q,\;
\lambda \in (0,1)
\right\},
\]
where \(c > 0\) is a hyperparameter that controls the scaling.

\bigskip
\noindent
Finally, \(\operatorname{MAUVE}(P, Q)\) is defined as the area under the divergence curve \(\mathcal{C}(P, Q)\). Its value lies between 0 and 100, with higher values indicating that \(Q\) is more similar to \(P\).

\subsection*{Coherence} Proposed by \citet{su2022contrastive}, the coherence metric is defined as the averaged log-likelihood of the generated text conditioned on the prompt as:

$$
\operatorname{Coherence}(\hat{\boldsymbol{x}}, \boldsymbol{x})=\frac{1}{|\hat{\boldsymbol{x}}|} \sum_{i=1}^{|\hat{\boldsymbol{x}}|} \log p_{\mathcal{M}}\left(\hat{\boldsymbol{x}}_i \mid\left[\boldsymbol{x}: \hat{\boldsymbol{x}}_{<i}\right]\right)
$$

where $\boldsymbol{x}$ and $\hat{\boldsymbol{x}}$ are the prompt and the generated text, respectively; [:] is the concatenation operation and $\mathcal{M}$ is the OPT model (2.7B) \cite{zhang2022opt}.






\subsection*{BERTScore} BERTScore has shown positive correlations with human judgements \cite{bert-score}, therefore, we propose using it as an additional metric for evaluating the quality of the generated texts by each generation method with respect to the corresponding reference texts provided by human experts. By using contextual embeddings, BERTScore \cite{bert-score} computes a similarity score for each token in the candidate sentence with each token in the reference sentence. 


\section{Human Evaluation}
\label{app:E}
To reflect the share of human raters favoring each decoding strategy, as depicted in Table \ref{tab:human_evaluation_double_exp} in the main paper, we apply the following scoring approach:

$$
\text { Score }_{\mathrm{ACS}}=\frac{\#(\mathrm{ACS} \text { is better })+0.5 \times \#(\text { tie })}{\#(\mathrm{ACS} \text { is better })+\#(\text { tie })+\#(\text { GUARD is better })} 
$$

$$
\text { Score }_{\mathrm{GUARD}}= 100 - \text { Score }_{\mathrm{ACS}}.
$$

\begin{table*}[h]
    \centering
\resizebox{1\textwidth}{!}{
\begin{tabular}{lcccccc}
\hline
\multirow{2}{*}{Dataset} & \multicolumn{3}{c}{Coherence} & \multicolumn{3}{c}{Fluency} \\

 & ACS is better & ACS and GUARD are similar & GUARD is better & ACS is better & ACS and GUARD are similar & GUARD is better \\
\hline
Wikitext & 19\% &  37\% & 44 \% & 11\% & 67\% & 23\% \\

Wikinews & 24\% & 33\% & 44\% & 12\% & 67\% & 22\% \\

BookCorpus & 36\% & 27\% & 38\% & 18\% & 69\% & 13\% \\
\hline
All & 26\% & 32\% & 42\% & 14\% & 67\% & 19\% \\
\hline

\end{tabular}
}
    \caption{Human evaluation results for ACS vs. GUARD across different datasets. Text generations are rated based on their semantic coherence and fluency.}
    \label{tab:human_evaluation_double_exp}
\end{table*}

Further, we measure the inter-rater agreement by computing Fleiss' kappa:

$$
\kappa_{Fleiss} = 0.41
$$
which reflects a moderate agreement across the human evaluators.

\begin{landscape}
    
\begin{figure*}[ht]
    \centering
    \adjustbox{rotate=0, max width= 1.6\textwidth, max totalheight= 1\textwidth}{%
        \includegraphics{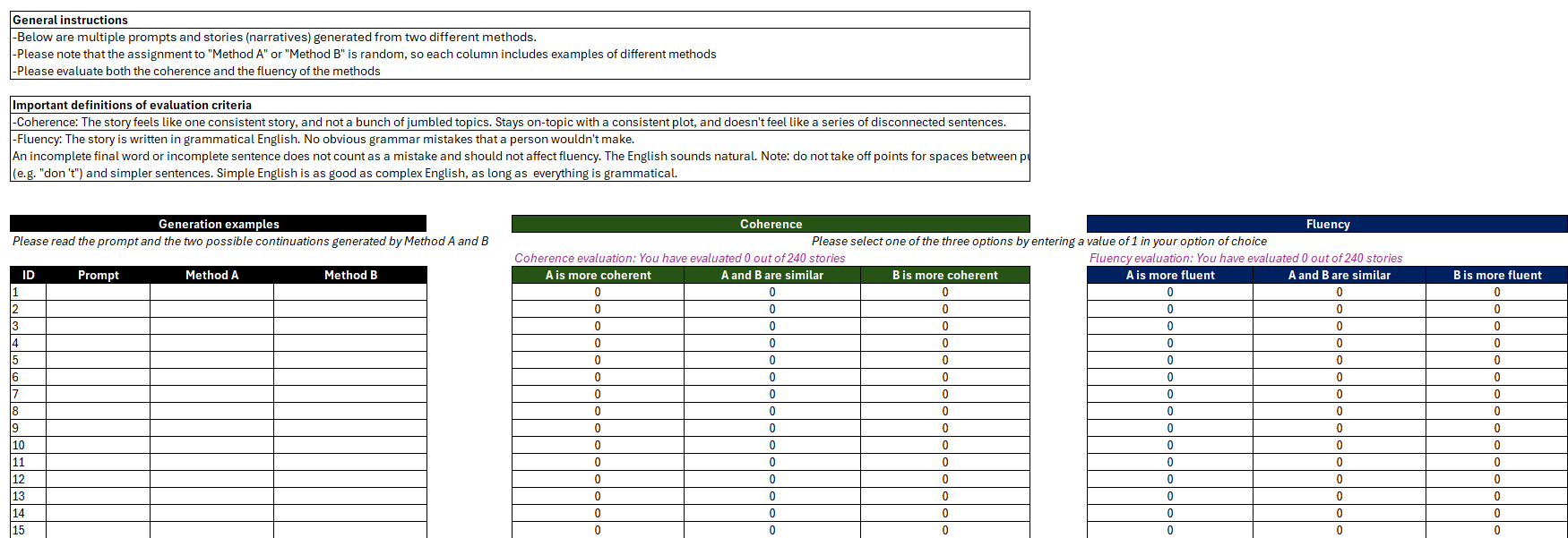}
    }
    \caption{Human evaluation form, including general instructions and definitions for the evaluation criteria.}
    \label{fig:human_evaluation_form}
\end{figure*}

\end{landscape}

\clearpage


\section{Prompt design for LLM-as-a-judge}
\label{app:F}

\begin{tcolorbox}[
    colback=gray!10,     
    colframe=black!70,   
    arc=8pt,             
    boxrule=1pt,         
    title=Instructions,  
    fonttitle=\bfseries, 
]

You are a language evaluation specialist tasked with conducting pairwise comparisons of machine-generated texts. You will compare outputs from two different decoding methods (A and B) that were generated using the same input prompt.
\subsubsection*{Evaluation Process}
For each text pair, assess which version demonstrates superior quality according to six specific metrics. Your evaluation should result in one of three judgments for each metric:
\begin{itemize}
    \item \textbf{Method A is better}
    \item \textbf{Both methods are similar}
    \item \textbf{Method B is better}
\end{itemize}
\paragraph{Important Note}
The underlying decoding methods are randomized in their labeling as ``Method A'' or ``Method B'' across different evaluations. Do not attempt to identify patterns based on these labels, as they are arbitrarily assigned for each comparison.
\subsubsection*{Evaluation Metrics}
\paragraph{1. Fluency}
 Measures how natural, smooth, and grammatically correct the text reads. Evaluates whether the language flows naturally without awkward phrasing, grammatical errors, or unnatural constructions.
\paragraph{2. Coherence}
Assesses logical connection between ideas and overall text organization. Evaluates whether the text maintains consistent themes, follows logical progression, and avoids contradictions or non-sequiturs.
\paragraph{3. Factuality}
 Measures accuracy and truthfulness of factual claims. Evaluates whether information presented is correct and free from errors, fabrications, or misrepresentations.
\paragraph{4. Informativeness}
 Assesses the substantive content and value of the information provided. Evaluates whether the text delivers meaningful, relevant content rather than being vague, repetitive, or content-poor.
\paragraph{5. Interestingness} 
Measures how engaging, compelling, or captivating the text is. Evaluates whether the content holds attention through creativity, unique insights, or engaging stylistic elements.
\paragraph{6. Story Development}
Assesses how effectively the narrative unfolds and progresses (where applicable). Evaluates character development, plot progression, pacing, and overall narrative structure in story-based texts.

\end{tcolorbox}

\clearpage


\section{Further Comparative Experiments}
\label{app:G}

\paragraph{Different Models} We examine the impact of different models on the quality of generated text, such as Llama-2, Gemma-7B, Llama-3.1, Mistral-v0.3, and Deepseek-llm-7B-base across three datasets: Wikinews, Wikitext, and BookCorpus. The results are detailed in Table \ref{tab:model_size_results} (diversity, MAUVE, and coherence) and Table \ref{tab:model_size_bertscore_results} (BERTScore) reveal notable differences between the models.

\begin{table}[H]
\centering
\resizebox{1\textwidth}{!}{
\begin{tabular}{llccccccccc}
\toprule
\multirow{2}{*}{Dataset} & \multirow{2}{*}{Model} & \multicolumn{3}{c}{Diversity} & \multicolumn{3}{c}{MAUVE} & \multicolumn{3}{c}{Coherence} \\
\cmidrule(lr){3-5} \cmidrule(lr){6-8} \cmidrule(lr){9-11}
 & & CS & ACS & GUARD & CS & ACS & GUARD & CS & ACS & GUARD  \\
\midrule
\multirow{4}{*}{Wikitext} 
 & Gemma-7B & 68.99 & 47.99 & \colorbox{Best}{\textbf{96.54}} & 85.05 & 68.45 & \colorbox{Best}{\textbf{88.16}} & -1.50 & -1.33 & -2.55 \\
 & Llama-3.1 & 9.0 & 81.17 & \colorbox{Best}{\textbf{95.84}} & 27.05 & 75.97 & \colorbox{Best}{\textbf{91.25}} & -0.74 & -1.52 & -2.55 \\
 & Mistral-v0.3 & \colorbox{Best}{\textbf{97.26}} & 89.82 & 96.59 & 85.19 & 81.59 & \colorbox{Best}{\textbf{90.58}} & -2.09 & -1.56 & -2.63 \\
 & Deepseek-base-7B & 77.43 & 32.14 & \colorbox{Best}{\textbf{96.55}} & 56.26 & 46.92 & \colorbox{Best}{\textbf{88.96}} & -2.49 & -1.23 & -2.66 \\
 \hline
 \multirow{4}{*}{Wikinews} 
 & Gemma-7B & 83.73 & 72.81 & \colorbox{Best}{\textbf{97.23}} & 87.26 & 86.34 & \colorbox{Best}{\textbf{90.76}} & -1.52 & -1.29 & -2.22 \\
 & Llama-3.1 & 46.84 & 89.66 & \colorbox{Best}{\textbf{96.45}} & 82.36 & 69.44 & \colorbox{Best}{\textbf{94.94}} & -1.25 & -1.66 & -2.20 \\
 & Mistral-v0.3 & \colorbox{Best}{\textbf{98.25}} & 93.42 & 96.85 & 83.89 & 80.75 & \colorbox{Best}{\textbf{90.49}} & -1.92 & -1.63 & -2.27 \\
 & Deepseek-base-7B & 94.13 & 61.96 & \colorbox{Best}{\textbf{97.14}} & 61.34 & 73.39 & \colorbox{Best}{\textbf{94.84}} & -3.69 & -1.66 & -2.23 \\
 \hline
\multirow{4}{*}{BookCorpus} 
 & Gemma-7B & 65.65 & 39.76 & \colorbox{Best}{\textbf{96.97}} & 86.65 & 70.59 & \colorbox{Best}{\textbf{91.29}} & -1.55 & -1.13 & -2.65 \\
 & Llama-3.1 & 16.45 & 84.86 & \colorbox{Best}{\textbf{96.21}} & 52.94 & 78.09 & \colorbox{Best}{\textbf{92.09}} & -1.00 & -1.64 & -2.50 \\
 & Mistral-v0.3 & \colorbox{Best}{\textbf{97.27}} & 90.74 & 96.78 & 79.03 & 75.50 & \colorbox{Best}{\textbf{94.56}} & -2.29 & -1.69 & -2.64 \\
 & Deepseek-base-7B & 62.33 & 42.61 & \colorbox{Best}{\textbf{96.86}} & 62.96 & 47.28 & \colorbox{Best}{\textbf{90.07}} & -2.41 & -1.24 & -2.61 \\
\bottomrule
\end{tabular}

}
\caption{Comparing GUARD with   CS ($k = 10, \alpha = 0.6$) and ACS ($q=1$)  across datasets and models of varying size. In Table \ref{tab:model_size_bertscore_results}, we also report the results of the influence of model architectures. Best results in \colorbox{Best}{\textbf{bold}}.}
\label{tab:model_size_results}
\end{table}

\noindent In Table \ref{tab:model_size_bertscore_results}, we also investigate the influence of model architectures on the performance of representative contrastive-search-based decoding methods.

\begin{table}[H]
\centering
\resizebox{1\textwidth}{!}{
\begin{tabular}{llccc}
\toprule
\multirow{2}{*}{Dataset} & \multirow{2}{*}{Model} & \multicolumn{1}{c}{CS} & \multicolumn{1}{c}{ACS} & \multicolumn{1}{c}{GUARD} \\
 & & BERTScore (\%) & BERTScore (\%) & BERTScore (\%)  \\
\midrule
\multirow{5}{*}{Wikitext} & Llama-2 &  81.13  & 80.59   & \colorbox{Best}{\textbf{81.50}} \\
 & Gemma-7B & \colorbox{Best}{\textbf{81.76}}  & 81.61 & 81.73 \\
 & Llama-3.1 & 80.89  & 80.98 & \colorbox{Best}{\textbf{81.76}} \\
 & Mistral-v0.3 & 81.14 & 81.25 & \colorbox{Best}{\textbf{81.84}} \\
 & Deepseek-base-7B & 80.00 & 80.02 & \colorbox{Best}{\textbf{81.30}} \\
 \hline
 \multirow{5}{*}{Wikinews} & Llama-2 & 82.00 & 82.23 & \colorbox{Best}{\textbf{83.39}} \\
 & Gemma-7B & \colorbox{Best}{\textbf{83.54}}  & 83.72 & 83.34 \\
 & Llama-3.1 & \colorbox{Best}{\textbf{83.89}} & 82.25 & 83.76 \\
 & Mistral-v0.3 & 82.76 & 83.08 & \colorbox{Best}{\textbf{83.52}} \\
 & Deepseek-base-7B & 77.28 & 80.03 & \colorbox{Best}{\textbf{83.32}}\\
 \hline
\multirow{5}{*}{BookCorpus} & Llama-2 & 79.74 & 81.36 & \colorbox{Best}{\textbf{81.92}} \\
 & Gemma-7B & 81.17  & 80.82 & \colorbox{Best}{\textbf{81.61}} \\
 & Llama-3.1 & 81.12 & 81.07 & \colorbox{Best}{\textbf{81.93}} \\
 & Mistral-v0.3 & \colorbox{Best}{\textbf{81.56}} & 81.43 & 81.83 \\
 & Deepseek-base-7B & 79.87 & 80.04 & \colorbox{Best}{\textbf{81.72}} \\
\bottomrule
\end{tabular}

}
\caption{Comparison of CS ($k = 10, \alpha = 0.6$), ACS ($q=1$), and GUARD in terms of BERTScore across different architectures/LLMs of varying sizes.  Best results in \colorbox{Best}{\textbf{bold}}.}
\label{tab:model_size_bertscore_results}
\end{table}

\clearpage

\paragraph{Hyperparameter Choice for Competing Decoding Strategies} We validate our choice of hyperparameters for the competing methods (Beam search, Top-$k$/Top-$p$ sampling) in Tables \ref{tab:different_beam_width}, \ref{tab:different_topk}, and \ref{tab:different_topp}.

\begin{table}[H]
\centering

\resizebox{1\textwidth}{!}{
\begin{tabular}{cccccccccc}
\toprule
\multirow{2}{*}{ Method } & \multicolumn{3}{c}{ Wikitext } & \multicolumn{3}{c}{ Wikinews } & \multicolumn{3}{c}{ BookCorpus } \\
\cmidrule(lr){2-4} \cmidrule(lr){5-7} \cmidrule(lr){8-10}
 & div. (\%) & MAUVE (\%) & coh. & div. (\%) & MAUVE (\%) & coh.  & div. (\%) & MAUVE (\%) & coh. \\
\midrule
$B$=3 & 15.24 & 32.26 & -0.81 & 25.78 & 58.02 & -0.91 & 5.48 & 20.23 & -0.59 \\
$B$=5 & 14.42 & 26.39 & -0.77 & 24.94 & 54.31 & -0.87 & 5.39 & 18.69 & -0.56\\
$B$=10 & 12.88 & 23.26 & -0.72 & 19.28 & 40.24 & -0.77 & 4.02 & 14.17 & -0.49\\
$B$=15 & 12.23 & 20.30 & -0.70 & 16.89 & 34.88 & -0.72 & 3.32 & 11.93 & -0.46 \\
$B$=20  & 11.52 & 17.63 & -0.67 & 15.45 & 34.71 & -0.69 & 2.94 & 10.68 & -0.44 \\
$B$=50  & 8.28 & 13.93 & -0.61 & 9.7 & 22.19 & -0.58 & 2.24 & 7.60 & -0.45 \\
\hline
 GUARD (Ours)  & 92.86 & 90.82 & -2.61 & 95.20 & 93.60 & -2.38 & 96.18 & 92.59 & -2.52\\

\bottomrule
\end{tabular}

}

\caption{Averaged automatic evaluation results for Qwen2.5-7B for beam search (with GUARD) with different beam-widths.}
\label{tab:different_beam_width}
\end{table}

\begin{table}[H]
\centering

\resizebox{1\textwidth}{!}{
\begin{tabular}{cccccccccc}
\toprule
\multirow{2}{*}{ Method } & \multicolumn{3}{c}{ Wikitext } & \multicolumn{3}{c}{ Wikinews } & \multicolumn{3}{c}{ BookCorpus } \\
\cmidrule(lr){2-4} \cmidrule(lr){5-7} \cmidrule(lr){8-10}
 & div. (\%) & MAUVE (\%) & coh. & div. (\%) & MAUVE (\%) & coh.  & div. (\%) & MAUVE (\%) & coh. \\
\midrule
$k$=3 & 59.01 & 82.40 & -1.33 & 74.21 & 91.46 & -1.38 & 47.18 & 87.86 & -1.44 \\
$k$=5 & 70.69 & 89.23 & -1.5 & 82.74 & 94.65 & -1.52 & 67.99 & 90.42 & -1.72\\
$k$=10 & 77.30 & 82.91 & -1.68 & 88.18 & 93.40 & -1.69 & 82.67 & 91.36 & -2.01\\
$k$=15 & 79.66 & 88.95 & -1.77 & 89.83 & 95.30 & -1.78 & 85.99 & 94.87 & -2.17 \\
$k$=20  & 81.62 & 89.73 & -1.84 & 90.17 & 94.11 & -1.84 & 88.14 & 92.96 & -2.56 \\
$k$=50  & 82.69 & 88.97 & -1.98 & 92.13 & 94.52 & -2.01 & 91.54 & 93.24 & -2.53 \\
\hline
 GUARD (Ours)  & 92.86 & 90.82 & -2.61 & 95.20 & 93.60 & -2.38 & 96.18 & 92.59 & -2.52\\

\bottomrule
\end{tabular}

}

\caption{Averaged automatic evaluation results for Qwen2.5-7B for Top-$k$ sampling (with GUARD) with different $k$.}
\label{tab:different_topk}
\end{table}

\begin{table}[H]
\centering
\resizebox{1\textwidth}{!}{
\begin{tabular}{cccccccccc}
\toprule
\multirow{2}{*}{ Method } & \multicolumn{3}{c}{ Wikitext } & \multicolumn{3}{c}{ Wikinews } & \multicolumn{3}{c}{ BookCorpus } \\
\cmidrule(lr){2-4} \cmidrule(lr){5-7} \cmidrule(lr){8-10}
 & div. (\%) & MAUVE (\%) & coh. & div. (\%) & MAUVE (\%) & coh.  & div. (\%) & MAUVE (\%) & coh. \\
\midrule
$p$=0.60 & 52.43 & 79.56 & -1.43 & 69.02 & 90.38 & -1.54 & 38.21 & 80.55 & -1.34 \\
$p$=0.70 & 64.30 & 83.92 & -1.60 & 77.56 & 93.74 & -1.65 & 56.81 & 85.18 & -1.58\\
$p$=0.80 & 73.80 & 83.80 & -1.77 & 84.98 & 96.08 & -1.83 & 72.88 & 90.73 & -1.81\\
$p$=0.90 & 79.72 & 88.62 & -1.97 & 89.96 & 96.67 & -2.02 & 84.76 & 93.30 & -2.12 \\
$p$=0.95  & 82.47 & 86.37 & -2.11 & 91.47 & 96.09 & -2.15 & 88.92 & 94.81 & -2.29 \\
\hline
 GUARD (Ours)  & 92.86 & 90.82 & -2.61 & 95.20 & 93.60 & -2.38 & 96.18 & 92.59 & -2.52\\

\bottomrule
\end{tabular}

}
\caption{Averaged automatic evaluation results for Qwen2.5-7B for Top-$p$ sampling (with GUARD) with different $p$.}
\label{tab:different_topp}
\end{table}

\section{Case Study: Generation Examples}
\label{app:H}

\begin{table}[H]
\small 
\centering
\begin{tabular}{|p{2cm}|p{13cm}|}
\hline
\textbf{Prompt} & \textit{Returning to Australia, Headlam became Deputy Chief of the Air Staff ( DCAS ) on 26 January 1965. He was appointed a Companion of the Order of} \\
\hline
\textbf{Reference} & the Bath ( CB ) " in recognition of distinguished service in the Borneo Territories " on 22 June. His tenure as DCAS coincided with the most significant rearmament program the Air Force had undertaken since World War II, and with manpower shortages stemming from this expansion and from Australia's increasing involvement in the security of South East Asia. \\
\hline
\textbf{CS} & St Michael and St George ( CMG ) in the 1966 New Year Honours. \textcolor{red}{What is the sentiment of this tweet? To determine the sentiment of this tweet, I will analyze the content and context of the information provided.} \\
\hline
\textbf{ACS} & St Michael and St George ( CMG ) on 26 June 1965. Promoted to Air Commodore on 1 July 1965, Headlam relinquished his appointment as DCAS on 15 July 1965. \textcolor{red}{He was appointed Air Officer Commanding RAAF Base Amberley on 15 July 1965}, a position he held until his retirement on 15 July 1967. \\
\hline
\textbf{GUARD} & the Bath ( CB ) in the 1965 New Year Honours. In May 1967, he moved to London as Chief Executive of the Civil Air Lines Branch, Department of Transport. Headlam became Air Commodore-in-Chief on 30 June. On 8 November, he assumed his third and last staff command when appointed Commander-in-Chief, Far East Region. His last appointment in July--August 1970 was the appointment as Chairman and Chief Executive Officer at Qantas Airways. \\
\hline
\end{tabular}
\captionof{table}{Text continuations based on prompts from the Wikitext dataset. Deviations from the context are highlighted in \textcolor{red}{red}. The results reveal that GUARD generates contextually relevant and fluent narratives, whereas the CS method deviates from the intended context and ACS exhibits repetitive patterns. When combined with the automatic evaluation results.}
\label{tab:case1}

\vspace{1em} 

\begin{tabular}{|p{2cm}|p{13cm}|}
\hline
\textbf{Prompt} & \textit{Two people have died as the result of a road traffic accident involving a truck and a car in Devon, England. The emergency services were called to the scene of} \\
\hline
\textbf{Reference} & a collision involving a Vauxhall Corsa and a DAF lorry at 0830 GMT on Saturday. Both occupants of the Vauxhall, a man and a woman, were pronounced dead at the scene of the collision, which was on the A361 road in the town of Knowstone, near South Molton. \\
\hline
\textbf{CS} & the accident, which took place on the A379 at 1.30 p.m. Police are appealing for witnesses to come forward and help with their investigation. Anyone who saw the accident or has any information is asked to contact Devon and Cornwall Police on 101, quoting log number 466 of 20/07/2019. \\
\hline
\textbf{ACS} & the accident at Bovey Tracey, Torquay, at 13:30 (14:30 CEST) on Saturday. Devon and Cornwall Police said the driver of the car, a 24-year-old man, was pronounced dead at the scene. \\
\hline
\textbf{GUARD} & the accident on 8 October, near Okehampton in north Devon. Both occupants of the car, which had collided with the back of the truck, have since died. \\
\hline
\end{tabular}
\captionof{table}{Text continuations based on prompts from the Wikinews dataset.}
\label{tab:case2}

\vspace{1em} 

\begin{tabular}{|p{2cm}|p{13cm}|}
\hline
\textbf{Prompt} & \textit{My eyes moved involuntarily from the pages of the book that I was no longer reading, to the faces of the girls who were now looking over at me. My}\\
\hline
\textbf{Reference} & heart began to pound in my chest, and I could hear the blood rushing in my ears. I could feel the heat of the color burning on my cheeks. I was embarrassed, no I was furious. \\
\hline
\textbf{CS} & heart skipped a beat as I realized that they were looking at me with a mix of curiosity and confusion. \\
\hline
\textbf{ACS} & eyes caught sight of one of the girls, and I couldn't help but stare. "Excuse me," one of the girls said, "but what are you looking at?" I looked up at her, and then back down at the book. \\
\hline
\textbf{GUARD} & heart beat faster as their gazes lingered upon me and, without even realising it myself, I was smiling back, my cheeks flushed and my eyes shining with happiness. \\
\hline
\end{tabular}
\captionof{table}{Text continuations based on prompts from the BookCorpus dataset.}
\label{tab:case3}
\end{table}

\end{document}